\documentclass[fleqn,usenatbib]{mnras}

\usepackage{newtxtext,newtxmath}

\usepackage[T1]{fontenc}

\DeclareRobustCommand{\VAN}[3]{#2}
\let\VANthebibliography\thebibliography
\def\thebibliography{\DeclareRobustCommand{\VAN}[3]{##3}\VANthebibliography}

\usepackage{graphicx}	
\usepackage{amsmath}	
\usepackage{color}
\usepackage{xcolor}

\usepackage{orcidlink}

\title[GraViT]{GraViT: Transfer Learning with Vision Transformers and MLP-Mixer for Strong Gravitational Lens Discovery}

\author[Parlange et al.]{
Ren\'e Parlange~\orcidlink{0000-0002-3900-7184},$^{1,2}$
Juan~C.~Cuevas-Tello~\orcidlink{0000-0002-7566-0412},$^{1}$
Octavio Valenzuela~\orcidlink{0000-0002-0523-5509},$^{2}$\thanks{E-mail: octavio@astro.unam.mx}
Omar~de~J.~Cabrera-Rosas~\orcidlink{0000-0001-8995-8933},$^{2}$\newauthor
Tom\'as Verdugo~\orcidlink{0000-0003-4062-6123},$^{3}$
Anupreeta More~\orcidlink{0000-0001-7714-7076},$^{4,5}$
and Anton~T.~Jaelani~\orcidlink{0000-0001-6282-5778}$^{6,7,8}$
\\
$^{1}$Facultad de Ingenier\'ia, Universidad Aut\'onoma de San Luis Potos\'i, 78290, San Luis Potos\'i, S.L.P., M\'exico\\
$^{2}$Universidad Nacional Aut\'onoma de M\'exico. Instituto de Astronom\'ia. A.P. 70-264, 04510. Ciudad de M\'exico, M\'exico\\
$^{3}$Universidad Nacional Aut\'onoma de M\'exico. Instituto de Astronom\'ia. A.P. 106, 22800. Ensenada, B.C., M\'exico\\
$^{4}$Inter-University Centre for Astronomy and Astrophysics (IUCAA), Post Bag 4, Ganeshkhind, Pune 411 007, India\\
$^{5}$Kavli IPMU (WPI), UTIAS, The University of Tokyo, Kashiwa, Chiba 277-8583, Japan\\
$^{6}$Astronomy Research Group and Bosscha Observatory, FMIPA, Institut Teknologi Bandung, Jl. Ganesha 10, Bandung 40132, Indonesia\\
$^{7}$U-CoE AI-VLB, Institut Teknologi Bandung, Jl. Ganesha 10, Bandung 40132, Indonesia\\
$^{8}$University Center of Excellence for Space Science, Technology and Innovation, Institut Teknologi Bandung, Jl. Ganesha 10, Bandung 40132, Indonesia\\
}

\date{}

\pubyear{2025}

\begin{document}
\label{firstpage}
\pagerange{\pageref{firstpage}--\pageref{lastpage}}
\maketitle

\begin{abstract}
Gravitational lensing offers a powerful probe into the properties of dark matter and is crucial to infer cosmological parameters. The Legacy Survey of Space and Time (LSST) is predicted to find \(\mathcal{O}(10^{5})\) gravitational lenses over the next decade, demanding automated classifiers. In this work, we introduce GraViT, a PyTorch pipeline for gravitational lens detection that leverages extensive pretraining of state-of-the-art Vision Transformer (ViT) models and MLP-Mixer. We assess the impact of transfer learning on classification performance by examining data quality (source and sample size), model architecture (selection and fine-tuning), training strategies (augmentation, normalization, and optimization), and ensemble predictions. This study reproduces the experiments in a previous systematic comparison of neural networks and provides insights into the detectability of strong gravitational lenses on that common test sample. We fine-tune ten architectures using datasets from HOLISMOKES VI and SuGOHI X, and benchmark them against convolutional baselines, discussing complexity and inference-time analysis. Our publicly available fine-tuned models provide a scalable transfer learning solution for gravitational lens finding in LSST.


\end{abstract}




\section{Introduction}
\label{sec:introduction}

Strong gravitational lensing, a phenomenon predicted by Einstein's theory of general relativity, occurs when a massive foreground object, such as a galaxy or galaxy cluster, bends and magnifies the light from a background source. This effect produces striking observational signatures, including multiple images, arcs, and Einstein rings, which provide a powerful tool for studying the distribution of dark matter \citep[see][and references therein]{Shajib2024SLGalaxies}, measuring cosmological parameters \citep{Birrer2024Time-DelayLensing, Pascale2024, Kelly2023}, and probing the evolution of galaxies \citep{Birrer2020}. The detection and analysis of strong lensing systems in large astronomical surveys remain challenging due to their rarity, the complexity of their morphologies, and the need for high-resolution imaging.

With the advent of large-scale sky surveys such as the Legacy Survey of Space and Time (LSST)\footnote{\url{https://www.lsst.org}} and the Euclid Space Telescope\footnote{\url{https://sci.esa.int/web/euclid}}, the volume of imaging data has grown exponentially, requiring automated methods to identify strong lensing candidates. Traditional approaches, such as visual inspection by experts or rule-based algorithms, are no longer scalable. Instead, machine learning techniques, particularly deep learning, have emerged as the state-of-the-art approaches for strong lens detection \citep{Jacobs2019a, Metcalf2019}. Among these, convolutional neural networks (CNNs) have dominated the field, leveraging their ability to capture local spatial features and hierarchical patterns in images \citep{Petrillo2019, Huang2020}. However, the recent rise of Vision Transformers (ViTs), which rely on self-attention mechanisms to model global relationships in data, has introduced a new paradigm in computer vision, raising the question of whether ViTs can outperform CNNs in the context of strong lensing.

CNNs have been highly successful in strong lens detection due to their inherent inductive biases, such as translation invariance and locality, which align well with the spatial structure of astronomical images. Models such as ResNet \citep{He2015DeepRecognition} and EfficientNet \citep{Tan2019EfficientNet:Networks} have achieved remarkable performance in identifying lensing features, such as arcs and rings, in survey data \citep{Jacobs2019b, Huang2021}. However, CNNs may struggle to capture long-range dependencies and global context, which are critical for distinguishing subtle lensing signals from complex backgrounds, such as overlapping galaxies or instrumental artifacts \citep{Shu2022HOLISMOKESProgram}. Recently, researchers  \citep{More2024SystematicLenses} conducted a systematic study of CNNs used for gravitational lens detection, showing that high performance is achievable on mid-size training sets, despite their focus on local features and limited contextual awareness.

In contrast, ViTs, which process images as sequences of patches and use self-attention to model relationships between all parts of an image, excel at capturing global context \citep{dosovitskiy2021imageworth16x16words}. This capability makes them particularly promising for tasks like strong lens detection \citep{ThuruthipillyFindingLensesBologna2021, Thuruthipilly2024}, where the lensing signal often spans a wide area of the image. Recent studies have demonstrated the potential of ViTs in astronomical applications, including galaxy classification \citep{Cao2024GalaxyCvT, Yao2021GalaxyTransformer} and
anomaly detection \citep{Hoyle2015}. However, ViTs typically require large amounts of training data and computational resources, which can be a limitation in astronomy, where labeled datasets are often small and imbalanced \citep{Rojas2022}. The lack of built-in inductive biases in ViTs means they can be less data-efficient, learning spatial relationships from scratch.

This study aims to compare the performance of ViTs and CNNs in the task of galaxy-galaxy strong gravitational lens detection. We evaluate both methods on the common test sample of real and simulated lenses, assessing their ability to identify lensing features, generalize to unseen data, and handle challenges such as noise and background contamination. By analyzing the strengths and limitations of each approach, we provide insights into their suitability for astronomical surveys and explore hybrid models that combine the local feature extraction of CNNs with the global context modeling of ViTs.

Our findings have implications not only for strong lens detection but also for the broader application of deep learning in astronomy. As the field moves toward increasingly complex and data-intensive observations, the choice of architecture will play a critical role in maximizing scientific returns. This work contributes to the growing body of research on machine learning in astronomy, offering a systematic evaluation of two leading paradigms and guiding the development of next-generation tools to explore the universe.

This paper is organized as follows. In \S\ref{sec:datasets}, we describe the strong lensing datasets, \S\ref{sec:architectures} introduces the architectures, and \S\ref{sec:experiments-models} details our transfer learning pipeline. Then, \S\ref{sec:results} reports performance metrics (AUC--ROC and F1 score), ensemble predictions, and inference-time analysis. Finally, \S\ref{sec:conclusions} presents our main conclusions and final remarks.

\section{Datasets}
\label{sec:datasets}

In this evaluation, we use datasets from the systematic comparison of neural networks~\citep{More2024SystematicLenses}, with image cutouts collected from the Hyper Suprime-Cam Subaru Strategic Program Public Data Release 2 (PDR2), where the GAMA09H field~\citep{Aihara2019SecondProgram} was reserved for the common test sample.

Certain test sets include gravitational lenses from the Kilo Degree Survey (KiDS) and Survey of Gravitationally-lensed Objects in HSC Imaging (SuGOHI)\footnote{\url{https://www-utap.phys.s.u-tokyo.ac.jp/~oguri/sugohi/}}. We refer the reader to these works for a rigorous description. Here, we briefly discuss the provenance of the datasets.

\subsection{Hyper Suprime-Cam}

The Hyper Suprime-Cam (HSC)\footnote{\url{https://subarutelescope.org/en/subaru2/instrument/hsc/}} is a digital camera for the 8.2 m Subaru Telescope, developed by the National Astronomical Observatory of Japan. The HSC-SSP Survey is an optical imaging survey conducted with the HSC, and its $1.77\,deg^2$ field-of-view optical camera~\citep{HSC_Public_Data_Release}.

\subsection{GAMA09H}

The Galaxy And Mass Assembly (GAMA)\footnote{\url{https://www.gama-survey.org/}} project is designed to exploit the latest generation of ground-based and space-borne survey facilities to study cosmology, galaxy formation, and evolution. 

HSC covers six discrete fields, named after overlapping regions from previous surveys, including GAMA09H \citep{Comparat2023TheeFEDS}.

\subsection{C21 training dataset}

\begin{figure}
    \centering    
    \includegraphics[width=0.8\columnwidth]{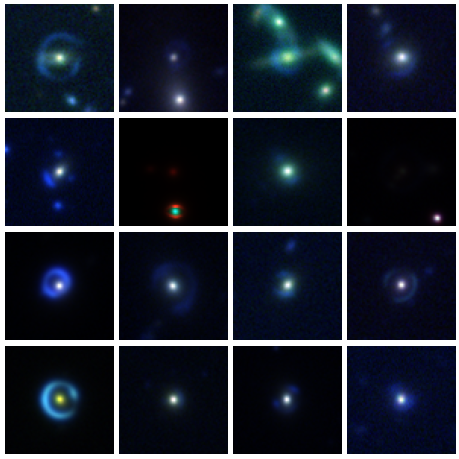}
    \caption{C21: mock lenses from \citet{Canameras2021HOLISMOKESSurvey}.}
    \label{fig:C21-mocklenses}
\end{figure}

\begin{figure}
    \centering    
    \includegraphics[width=0.8\columnwidth]{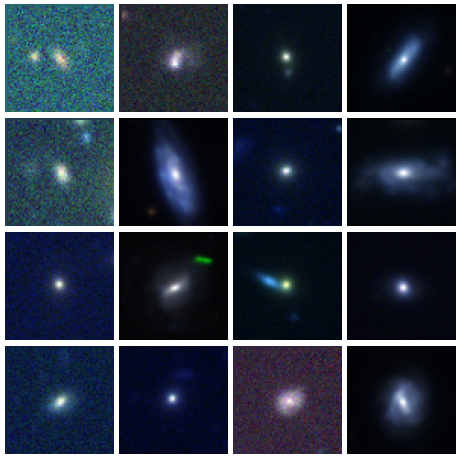}
    \caption{C21: non-lenses from \citet{Canameras2021HOLISMOKESSurvey}.}
    \label{fig:C21-nonlenses}
\end{figure}

C21 training set from HOLISMOKES\footnote{\url{https://shsuyu.github.io/HOLISMOKES/site/}} VI ~\citep{Canameras2021HOLISMOKESSurvey} was stored using the Flexible Image Transport System (FITS), as data cubes of $72\times72$ pixels with \textit{gri} photometric bands. The \texttt{astropy} library \citep{astropy:2013,astropy:2018,astropy:2022}  is used for data loading. Its bands are mapped: $(gri \rightarrow BGR$), or in RGB order ($irg \rightarrow RGB$). This mapping aligns the bands by increasing wavelength to match the spectral order expected by pre-trained models.

The dataset has 40,000 lenses and 40,000 non-lenses (plus 500 samples for validation of each class), with a random sample displayed in Figures \ref{fig:C21-mocklenses} and \ref{fig:C21-nonlenses}. It was obtained from random sky positions to limit biases from small-scale seeing and depth variations. The GAMA09H field was reserved for the common test sample \citep{More2024SystematicLenses}. 

As positive examples, they produced realistic galaxy-scale lens simulations with synthetically lensed arcs on HSC \textit{gri} images of Luminous Red Galaxies (LRGs). This accounts for the quality of HSC imaging and for the presence of artifacts and neighboring galaxies.

They modeled lens mass distributions as Singular Isothermal Ellipsoids (SIE) using LRG redshifts from the Sloan Digital Sky Survey (SDSS), velocity dispersions, and light profiles, adding external shear to create a uniform Einstein radius range (0.75$''$--2.5$''$) that captures wider separations and fainter ($z > 0.7$) lenses. 

High signal-to-noise background galaxies from the Hubble Ultra Deep Field (HUDF) were lensed with \texttt{GLEE}  \citep{Suyu2010TheSL2SJ085440121, Suyu2012DISENTANGLINGB1933+503}, convolved with HSC PSFs, and only mocks meeting strict magnification ($\mu \ge 5$), S/N ($>5$), and brightness criteria were accepted, including both quad and double images.

\subsection{J24 training dataset}

\begin{figure}
    \centering    
    \includegraphics[width=0.8\columnwidth]{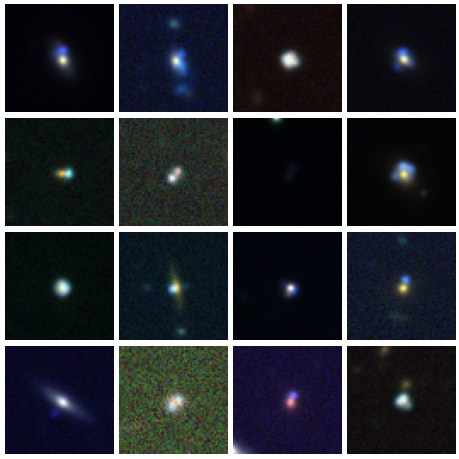}
    \caption{J24: mock lenses from \citet{Jaelani2024SurveyNetworks}.}
    \label{fig:J24-mocklenses}
\end{figure}

\begin{figure}
    \centering        \includegraphics[width=0.8\columnwidth]{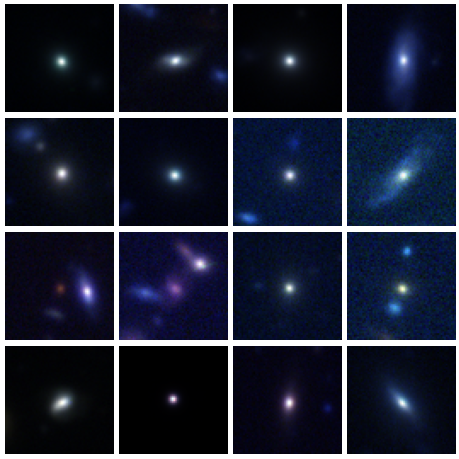}
    \caption{J24: non-lenses from \citet{Jaelani2024SurveyNetworks}.}
    \label{fig:J24-nonlenses}
\end{figure}

J24 \citep{Jaelani2024SurveyNetworks} training set has 18,660 lenses and 18,660 non-lenses (with no validation set) from  SuGOHI X, stored in FITS data cubes. The images are 64 x 64 pixels, using \textit{gri} bands as well. A small random sample of both classes is shown in Figures \ref{fig:J24-mocklenses} and \ref{fig:J24-nonlenses}.

The authors use data from the HSC-SSP Public Release 2 (PDR2) \citep{Aihara2019SecondProgram}. The HSC-SSP Survey footprint overlaps with SDSS, therefore they collected spectroscopic redshifts from the catalogs of the SDSS Data Release 16. The distribution of lensing properties in J24 was constructed using \texttt{SIMCT}\footnote{\url{https://github.com/anupreeta27/SIMCT}} \citep{More2016SpaceWarpsScience}.

Their pipeline adds simulated lensed galaxies around massive galaxies from the parent catalog. Lensing galaxies are modeled with SIE profile and external shear. The mass model parameters are generated by using the magnitudes, photometric redshift and ellipticity of the galaxies. Background lensed galaxies are assumed to follow a Sérsic profile, and the sources are drawn from galaxy luminosity function, while colors are taken from galaxies detected in data from the CFHTLenS catalog \citep{Hildebrandt2012CFHTLenS:Photometry, Erben2013CFHTLenS:Products}.

A detailed comparison of the methodologies\footnote{S22 \citep{Shu2022HOLISMOKESProgram} was unavailable and I24 \citep{ishida24} used J24.} used to generate C21 and J24 training datasets can be found in Table \ref{tab:comparison} of the Appendix.

\subsection{Common test sample}

The common test sample \citep{More2024SystematicLenses}  consists of 12 strong lensing test datasets designed to benchmark neural networks. The \textit{gri} bands are stored jointly in a single data cube per sample, with resolutions of $64\times64$ or $101\times101$ px. These datasets include known galaxy-scale lenses (L1), lens candidates identified by their networks (L2), and mock lenses from \citet{Canameras2021HOLISMOKESSurvey} (L3) and \citet{Jaelani2024SurveyNetworks} (L4). Additionally, the sample contains various types of non-lenses: random non-lenses (N1), non-lenses selected following \citet{Canameras2021HOLISMOKESSurvey} (N2), \citet{Shu2022HOLISMOKESProgram} (N3), and \citet{Jaelani2024SurveyNetworks} (N4), as well as a subset of challenging non-lenses (N5).

\begin{table}
\centering
\caption{Common test sample: composition, counts, and resolution.}
\begin{tabular}{c c c r r r c}
\hline
\textbf{ID} & \textbf{Lens} & \textbf{Non-Lens} & \textbf{L} & \textbf{N} & \textbf{Total} & \textbf{Resolution (px)} \\
\hline
a & L$_1$\,L$_2$ & N$_1$          & 181   & 3\,000  & 3\,181  & 64 \\
b & L$_1$\,L$_2$ & N$_2$          & 181   & 3\,000  & 3\,181  & 64 \\
c & L$_1$\,L$_2$ & N$_3$          & 181   & 3\,000  & 3\,181  & 64 \\
d & L$_1$\,L$_2$ & N$_4$          & 181   & 3\,000  & 3\,181  & 64/101 \\
e & L$_1$\,L$_2$ & N$_5$          & 181   &   730   &   911   & 64 \\
f & L$_1$\,L$_2$ & N$_1$--N$_5$   & 181   & 12\,730 & 12\,911 & 64/101 \\
g & L$_3$        & N$_2$          & 3\,000 & 3\,000  & 6\,000  & 64 \\
h & L$_3$        & N$_3$          & 3\,000 & 3\,000  & 6\,000  & 64 \\
i & L$_3$        & N$_4$          & 3\,000 & 3\,000  & 6\,000  & 64/101 \\
j & L$_4$        & N$_2$          & 3\,000 & 3\,000  & 6\,000  & 64/101 \\
k & L$_4$        & N$_4$          & 3\,000 & 3\,000  & 6\,000  & 101 \\
l & L$_1$--L$_4$ & N$_1$--N$_5$   & 6\,181 & 12\,730 & 18\,911 & 64/101 \\
\hline
\end{tabular}
\label{tab:common_test_minimal}
\end{table}

\begin{figure*}
    \centering        \includegraphics[width=0.9\textwidth]{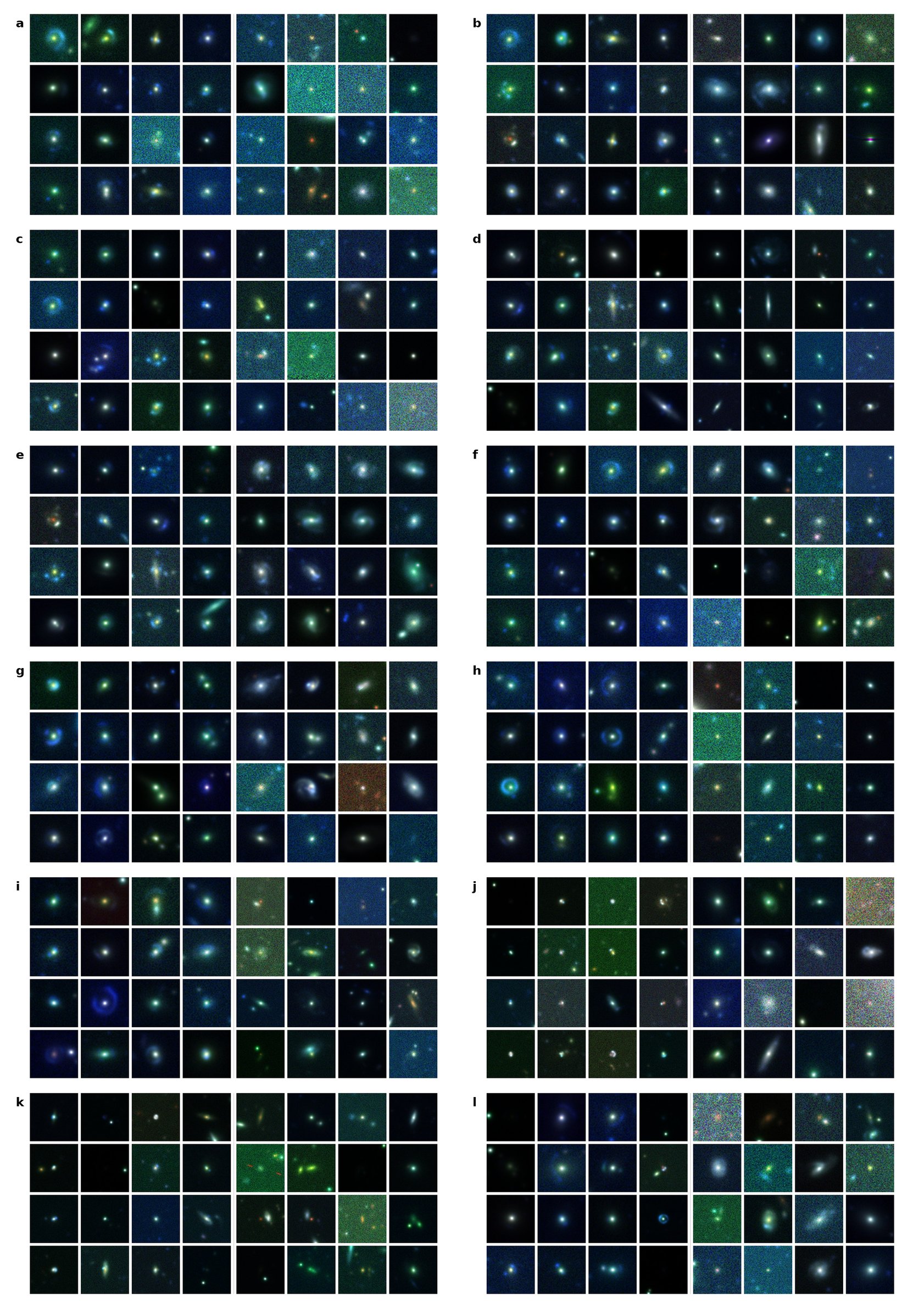}
    \caption{Common test sample \citep{More2024SystematicLenses}. Panels: left (lenses) vs. right (non-lenses)  for each test set labeled \textit{a} to \textit{l}.}
    \label{fig:common-test-sample}
\end{figure*}

\section{Neural Architectures}
\label{sec:architectures}

This work replicates and extends the experiments in the systematic comparison of neural networks \citep{More2024SystematicLenses}, evaluating the ViT, and nine variants, including hybrid models that introduce convolutions, and the MLP-Mixer, which instead uses multilayer perceptrons to mix information across tokens and channels, and does not rely on convolutional networks or the self-attention mechanism.

\subsection{Systematic Comparison of Neural Networks}

The gravitational lens classifier in \citet{Canameras2021HOLISMOKESSurvey} (C21) is a residual neural network, ResNet-18 architecture \citep{He2015DeepRecognition}. It was trained and validated on \textit{gri} images of the HSC Survey, augmented with small random shifts ranging between $-5$ and $+5$ pixels, and square root stretch after clipping negative pixels to zero. Their optimizer uses mini-batch gradient descent with a binary cross-entropy loss function, a batch size of 128, a learning rate of $6 \times 10^{-4}$, a weight decay of $10^{-3}$, and a momentum fixed at 0.9. The network sets an early stopping mechanism that saves the best model based on the minimum validation loss.

The lens classifier in J24 \citep{Jaelani2024SurveyNetworks} uses a CNN inspired by the architecture used in \citet{Jacobs2017FindingNetworks}. It was trained on HSC images with \textit{gri} bands, using data augmentation (random rotation, flipping, resizing and channel shift), and  the Adam optimizer to minimize the cross-entropy loss with a learning rate of  $5 \times 10^{-5}$. The CNN was trained for 52 epochs (with a maximum of 100) using mini-batch stochastic gradient descent with 128 images per batch and early stopping after a patience of 5 epochs for improvement in accuracy or loss. The parent sample of 2.3 million galaxies was selected based on the criteria of multi-band magnitudes, stellar mass, star formation rate, extendedness limit, and photometric redshift range.

The strong lens classifier I24 \citep{ishida24} uses a CNN architecture and has no dataset, therefore, used J24. They transformed the FITS data cubes using SDSS normalization \citep{Lupton2004PreparingData}. The data augmentation is applied directly to the input training and validation data and it does not create multiple objects. The network uses early stopping to save the best model with a patience of 5 epochs.

\subsection{Vision Transformers}

\begin{figure*}
    \centering        \includegraphics[width=0.8\textwidth]{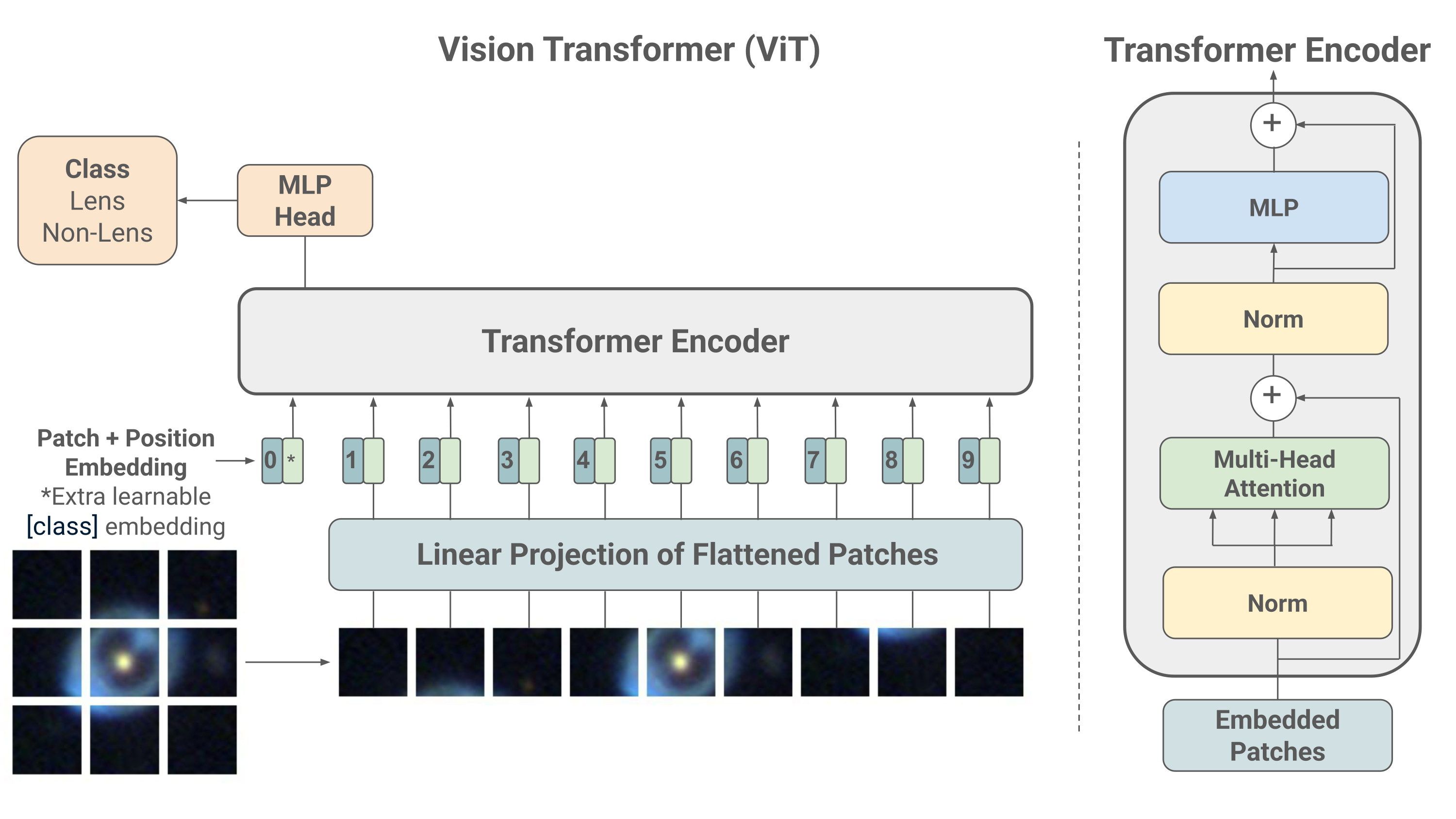}
    \caption{Gravitational lens detection with the Vision Transformer (ViT) architecture \citep{dosovitskiy2021imageworth16x16words}.}
    \label{fig:vit}
\end{figure*}

The transformer architecture originated in natural language processing \citep{DBLP:journals/corr/VaswaniSPUJGKP17}, was adapted to computer vision in the Image Transformer \citep{Parmar2018}, and was later established for image classification with the Vision Transformer (ViT) \citep{dosovitskiy2021imageworth16x16words}, depicted in Figure~\ref{fig:vit}.
The ViT and its variants have been widely adopted in astronomy for many applications. For instance, the morphological classification of galaxies using the Convolutional Vision Transformer (CVT) \citep{Cao2024GalaxyCvT}, and an efficient ViT \citep{Yao2021GalaxyTransformer} based on Linformer \citep{Wang2020Linformer:Complexity} that approximates self-attention, which has $\mathcal{O}(n^2)$ complexity, as a low-rank matrix with linear complexity $\mathcal{O}(n)$. Moreover, the detection transformer (DETR) \citep{Carion2020}, recently adopted in gravitational lensing studies, enabled end-to-end detection of strongly lensed arcs in galaxy clusters \citep{Jia_2023}.

Beyond classification, the ViT architecture has also been used in gravitational lens modeling \citep{Huang2023StrongTransformer}, a multi-regression task of parameter and uncertainty estimation. They had eight target parameters to be predicted: the Einstein radius $\theta_E$, the ellipticities $e_1$ and $e_2$, the radial power-law slope $\gamma_0$, the coordinates of the mass center $\theta_1$ and $\theta_2$, the effective radius $R_{\text{eff}}$, and the Sérsic index $n_{\text{Sérsic}}$.

In the context of LSST, the ViT is part of TEGLIE \citep{Grespan2024TEGLIE:KiDS}, a framework for strong gravitational lens detection that adopts fine-tuning on augmentations of real observations, to bridge the gap between simulated lenses and surveys. 

As reported in a recent study, there was a lens search in the Dark Energy Survey (DES) with Space Warps \citep{gonzalez2025discoveringstronggravitationallenses} where the authors proposed a pre-trained ViT to classify and reduce 236 million objects in DES to 22,564 targets of interest, then inspected by citizen scientists, who ruled out $90\%$ as false positives, effectively sifting through the candidate list.

Our pipeline is based on the Torch Image Models (\texttt{timm}\footnote{\url{https://huggingface.co/timm}}) library \citep{rw2019timm}. In particular, we fine-tune the classification head or part of the architecture (e.g., half of the layers, blocks, stages, or all of them) in models that were pre-trained on ImageNet-1k \citep{Deng2009ImageNet:Database} and ImageNet-21k \citep{Ridnik2021ImageNet-21KMasses}.

This study explores the Vision Transformer (ViT) and nine architectures that extend or refine its framework for visual recognition.

\subsubsection*{Vision Transformer (ViT)}
ViT \citep{dosovitskiy2021imageworth16x16words} replaces convolutions with a Transformer architecture by dividing an image into patches and processing them as tokens. This enables long-range dependencies and contextual understanding but requires extensive datasets for optimal performance. Data augmentation and regularization strategies are needed to improve its generalization ability.

\subsubsection*{Data-efficient image Transformer (DeiT)}
DeiT \citep{Touvron2020TrainingAttention} optimizes the training procedure for Transformers on smaller datasets by introducing a knowledge distillation mechanism with a dedicated distillation token. This allows DeiT to match or surpass the accuracy of convolutional neural networks, while improving computational efficiency.

\subsubsection*{Class-Attention in Image Transformers (CaiT)}
CaiT \citep{Touvron2021GoingDW} refines the ViT framework with deeper self-attention layers and class-attention mechanisms. These modifications provide stability during training and improve performance by explicitly separating patch-wise and classification attention mechanisms, leading to better feature aggregation.

\subsubsection*{DeiT III}
Building upon DeiT architecture, DeiT III \citep{Touvron2022DeiTViT} further improves training strategies, including stronger regularization and augmentation techniques. These adaptations enable effective training of ViTs on mid-sized datasets, narrowing the performance divide between data-intensive Transformers and traditional convolutional models.

\subsubsection*{Swin Transformer}
Swin Transformer \citep{Liu2021SwinWindows} tackles the quadratic complexity of standard self-attention by introducing a hierarchical architecture with shifted windows. The shift operation enables the model to capture relationships between adjacent windows and aggregate information across the entire image to build a global context. This design allows multi-scale feature extraction for image classification.

\subsubsection*{Twins-SVT}
Twins-SVT \citep{Chu2021Twins:Transformers} introduces spatially separable self-attention to reduce computational cost while preserving global feature interactions. It balances efficiency and representation power, providing a strong backbone for vision models.

\subsubsection*{Twins-PCPVT}
Twins-PCPVT \citep{Chu2021Twins:Transformers} incorporates a pyramid structure similar to CNNs, enabling effective multi-scale feature representation. This architecture has the ability to capture fine-grained and global patterns simultaneously, improving both classification and object detection.

\subsubsection*{Pooling-based Vision Transformer (PiT)}
PiT \citep{Heo2021RethinkingTransformers} integrates spatial pooling within the Transformer framework, reducing the number of tokens at deeper layers. This mimics the hierarchical nature of CNNs, improving efficiency while maintaining high performance in feature representation.

\subsubsection*{Convolutional Vision Transformer (CvT)}
CvT \citep{Wu2021CvT:Transformers} introduces convolutional token embeddings and projections within the Transformer architecture. By incorporating convolutional layers, CvT improves local feature extraction, spatial hierarchy, and computational efficiency while retaining the benefits of global attention from Vision Transformers.

\subsection{Multilayer Perceptron}
Multilayer perceptrons (MLPs), also known as backpropagation networks~\citep{rumelhart1986learning}, are feedforward neural networks composed of stacked, fully-connected layers that apply affine transformations followed by nonlinear activation functions.

\subsubsection*{MLP-Mixer}

MLP-Mixer \citep{Tolstikhin2021MLP-Mixer:Vision} is an innovative architecture that relies exclusively on MLPs instead of convolutions or self-attention, to separately mix information across tokens (spatial patches) and channels (e.g., the \textit{gri} photometric bands), jointly modeling spatial and spectral representations. 

\begin{figure*}
    \centering        \includegraphics[width=0.8\textwidth]{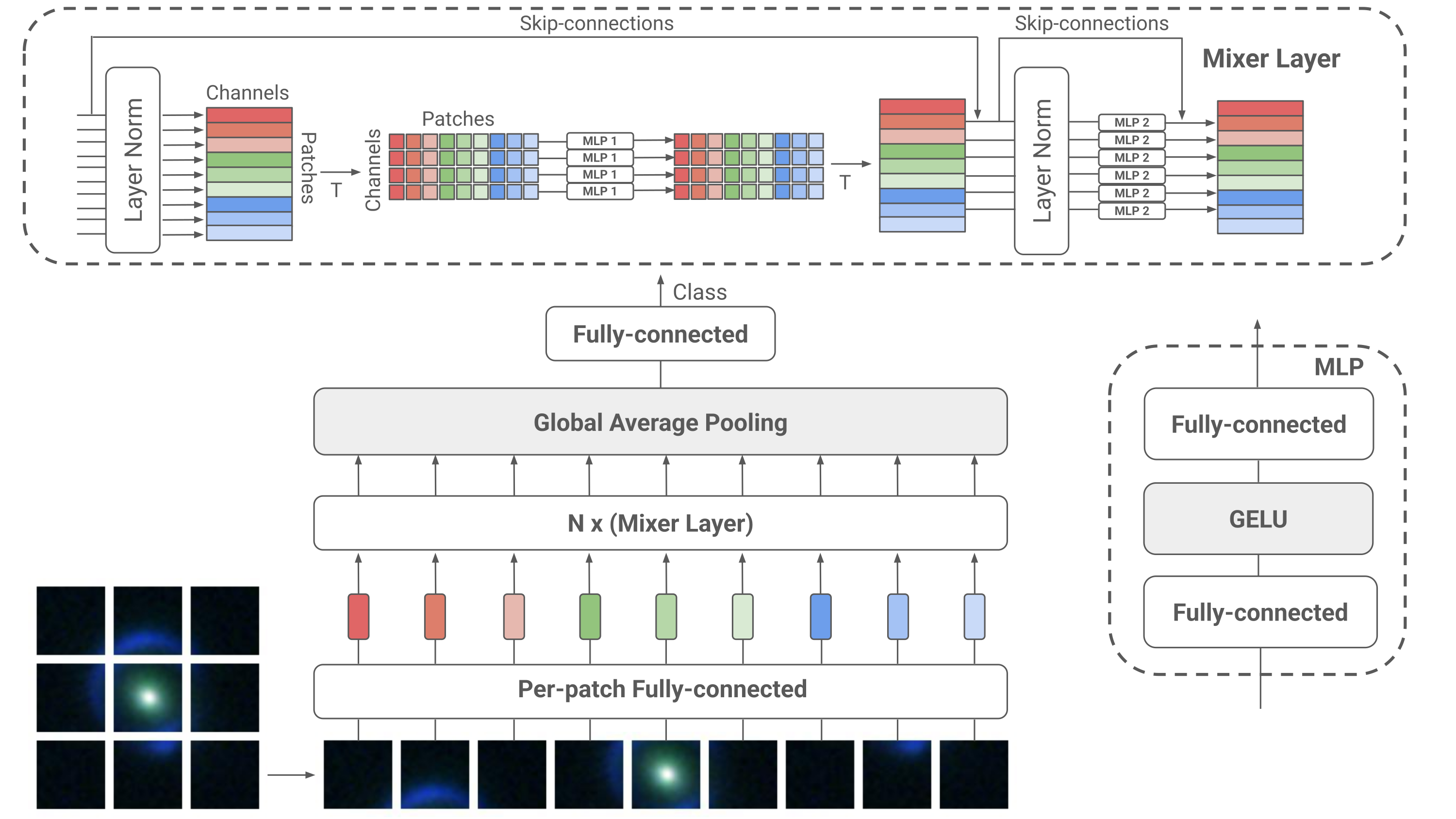}
    \caption{Lens classification diagram based on the MLP-Mixer architecture \citep{Tolstikhin2021MLP-Mixer:Vision}.}
    \label{fig:mlp-mixer}
\end{figure*}

It splits images into fixed-size patches, linearly projects them into a tokens$\times$channels matrix, and alternates between \textit{token-mixing} MLPs (which aggregate spatial information across patches) and \textit{channel-mixing} MLPs (which blend feature channels within each patch), efficiently capturing both local and global patterns (see Figure~\ref{fig:mlp-mixer}).

Its operations are strictly limited to matrix multiplications, tensor reshapes and transpositions  (i.e., no positional embeddings), nonlinear activation functions, residual connections \citep{He2015DeepRecognition}, dropout \citep{srivastava2014dropout}, and layer normalization \citep{BaLayerNormalization}. A final global pooling step then averages the features from all patches, which in turn produces a single fixed-length vector that is insensitive to the input patch permutations. 

In contrast, convolutional neural networks encode structure through localized, weight-shared filter kernels, producing representations that are by nature equivariant to translations but sensitive to local spatial ordering of features.

\section{Experiments and Models}
\label{sec:experiments-models}

As presented in the previous sections, we evaluate 10 neural architectures: ViT, DeiT, CaiT, DeiT III, Swin, Twins-SVT, Twins-PCPVT, PiT, CvT, and MLP-Mixer. 

We refer to these trained instances as models, resulting in a total of ten different models for each dataset and transfer learning setting, where \textit{transfer learning} \citep[see][and references therein]{Steiner2021HowTT} denotes the retraining of a pre-trained model for a downstream task (e.g., fine-tuning) with fewer training samples.

Following \citet{More2024SystematicLenses}, we reconstruct the different combinations of test datasets (\textit{a} to \textit{l}) with lenses (L1, L2, L3, L4) and non-lenses (N1, N2, N3, N4, N5), see Table \ref{tab:aucroc_integrated}. For each test set, we conduct a distinct evaluation based on the training data and the degree of fine-tuning. The symbols represent the experiments as follows:
\begin{itemize}
\item $\mathcal{A}$ corresponds to data from C21 \citep{Canameras2021HOLISMOKESSurvey},
\item $\mathcal{B}$ corresponds to data from J24 \citep{Jaelani2024SurveyNetworks},
\item $\mathcal{C}$ represents the combination of both datasets (C21+J24).
\end{itemize}
Additionally, a reduced version of C21, denoted as $\mathcal{S}$, contains 18,660 samples per class instead of the original 40,000, to match the sample size of J24.
The subscripts indicate the fine-tuning strategy applied:
\begin{itemize}
\item $\mathcal{X}_1$ refers to fine-tuning only the classification head,
\item $\mathcal{X}_2$ corresponds to fine-tuning half of the architecture,
\item $\mathcal{X}_3$ represents fine-tuning the entire architecture,
\end{itemize}
where $\mathcal{X}\in\{\mathcal{A}, \mathcal{B}, \mathcal{C}, \mathcal{S}\}$.

For each dataset and fine-tuning setting, the ten models were trained for up to $100$ epochs, with early stopping triggered if there is no improvement in validation loss for $20$ epochs.

Our \texttt{PyTorch} pipeline uses mixed precision with \texttt{autocast}, \texttt{GradScaler} for numeric stability in training, and multi-GPU execution via \texttt{DataParallel}. However, it is now recommended to use \texttt{DistributedDataParallel} for both single-node and multi-node training due to its increased efficiency and scalability.

We use the \texttt{AdamW} optimizer \citep{Loshchilov2017DecoupledRegularization} with a learning rate (LR) of \(10^{-4}\) and a decoupled weight decay of \(10^{-2}\). A scheduler to \texttt{ReduceLROnPlateau} monitors validation loss, reducing LR by a factor of 0.1 after 5 epochs without improvement, enabling fine-tuning without drastic oscillations. Unlike \texttt{Adam} \citep{Kingma2014Adam:Optimization}, \texttt{AdamW} applies weight decay separately from the optimization step, improving generalization by limiting the influence of L2 regularization on weight updates.

\begin{figure}
    \centering        
    \includegraphics[width=1.0\columnwidth]{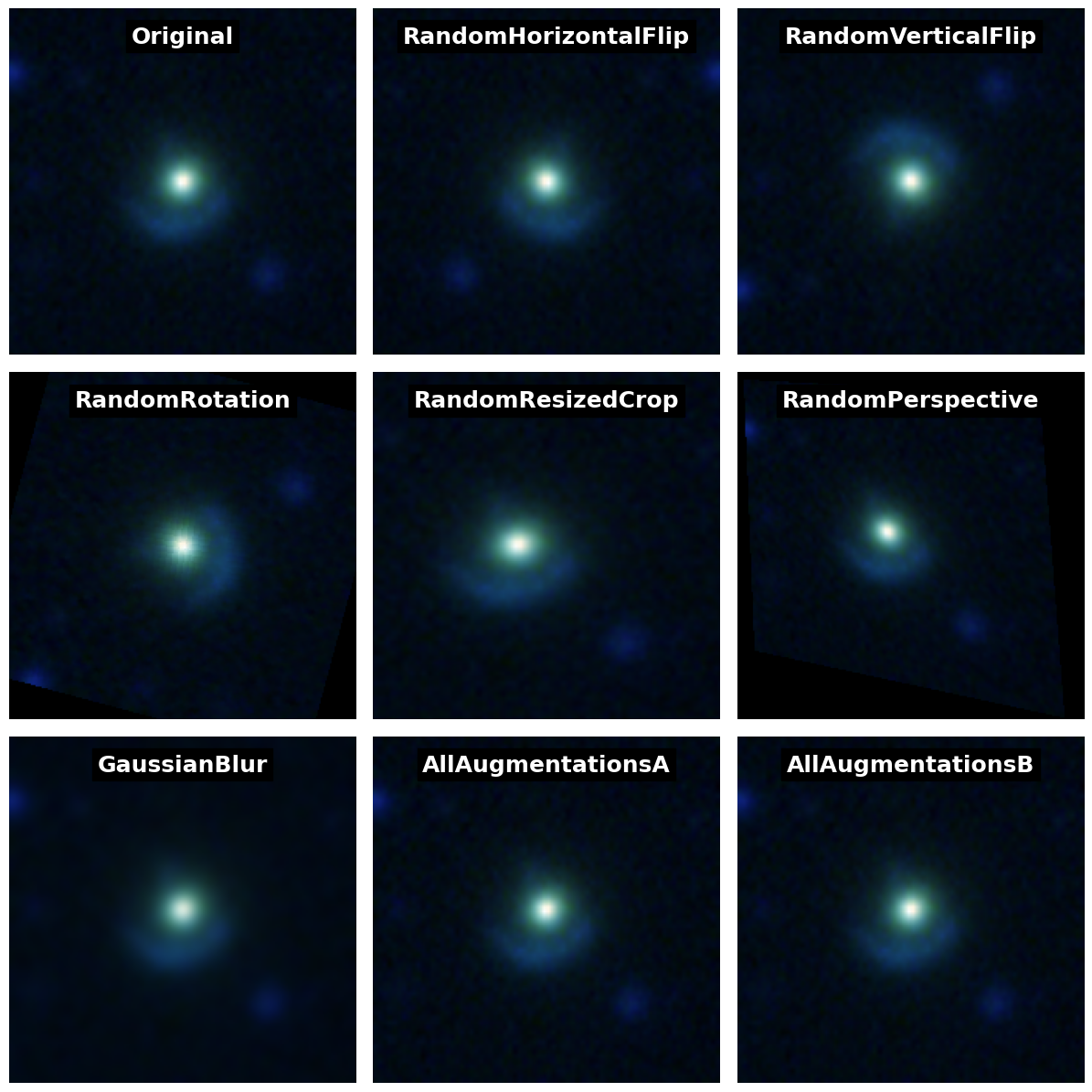}
    \caption{Example of augmented lens from C21 \citep{Canameras2021HOLISMOKESSurvey}. Data augmentations applied to the training data loader are listed on Table \ref{tab:train_aug_norm}.}
    \label{fig:augmentations}
\end{figure}

Stochastic depth \citep{Huang2016DeepDepth} randomly skips layers at a rate of \(0.1\), with a 10\% probability of being excluded during training. Unlike dropout \citep{srivastava2014dropout}, which randomly deactivates individual neurons, stochastic depth selectively drops entire layers or pathways. This regularization technique mitigates overfitting, stabilizes the training of deep networks, and encourages diverse feature representations, as well as smoother gradient flow.

The training data undergoes transformations such as normalization, as well as the six data augmentations described in Table \ref{tab:train_aug_norm} and depicted in Figure~\ref{fig:augmentations}. The validation data is not augmented, just resized, and normalized with ImageNet \citep{Deng2009ImageNet:Database} statistics:

\begin{equation}
\text{Normalization}(x) = \frac{x - \mu}{\sigma}
\end{equation}

where \(\mu\) and \(\sigma\) represent the channel-wise mean and standard deviation values in RGB order:
\begin{equation}
\mu = [0.485, 0.456, 0.406], \quad \sigma = [0.229, 0.224, 0.225]
\end{equation}

\begin{table}
\centering
\caption{Transformations of our training data loader: data augmentation and normalization. Augmentations are applied with a 50\% probability.}
\label{tab:train_aug_norm}
\resizebox{\columnwidth}{!}{%
\begin{tabular}{lll}
\hline
\textbf{Technique}      & \textbf{Description}                                    & \textbf{Type}         \\ \hline
Resize                  & Resize to 224$\times$224 pixels                        & Transformation        \\ 
Horizontal Flip         & 50\% chance to flip horizontally                        & Augmentation          \\ 
Vertical Flip           & 50\% chance to flip vertically                          & Augmentation          \\ 
Rotation                & Rotate within $[-30^\circ, 30^\circ]$                   & Augmentation          \\ 
Resized Crop            & Random crop with scale [0.8, 1.2]                        & Augmentation          \\ 
Perspective             & Perspective transform (distortion scale 0.4)       & Augmentation          \\ 
Gaussian Blur           & Blur with 5$\times$5 kernel, $\sigma$ [0.1, 2.0]            & Augmentation          \\ 
ToTensor                & Convert image to tensor                                  & Transformation        \\ 
Normalize               & Normalize using ImageNet statistics                          & Transformation        \\ \hline
\end{tabular}%
}
\end{table}

This is standard image normalization for models pre-trained on ImageNet-1k with 1,000 classes (\textit{ft\_in1k}). Note that the ViT and MLP-Mixer by Google are pre-trained beforehand on its superset, ImageNet-21k \citep{Ridnik2021ImageNet-21KMasses} with 21,000 classes and then fine-tuned on ImageNet-1k (1,000 classes), referred to as \textit{in21k\_ft\_in1k} in the \texttt{timm} library, following the same normalization procedure.

The pre-trained architectures are fine-tuned using datasets C21, J24, and their combination (C21+J24), with varying degrees of fine-tuning: unfreezing only the classification head, fine-tuning half of the layers, or updating all layers, blocks, or stages in the architecture.

\subsection{Metrics}

The pipeline provides detailed logging of training metrics, tracking the progression of loss, accuracy, the Area Under the Receiver Operating Characteristic Curve (AUC--ROC) and the F1 score.

The checkpoints save the best model based on validation loss. The results of the evaluation are saved in a Pandas DataFrame and stored in a CSV file for further analysis. 

All the confusion matrices and aggregated ROC curves for each experiment can be found in Jupyter notebooks at the repository\footnote{\url{https://github.com/parlange/gravit}}.

We evaluate our transformer-based models using AUC--ROC and F1 score, in direct comparison with the results of \citet{More2024SystematicLenses} for a consistent benchmark.

These metrics are defined as follows:

\begin{equation}
    \text{Precision} = \frac{TP}{TP + FP},
\end{equation}

\begin{equation}
    \text{Recall (Sensitivity)} = \frac{TP}{TP + FN},
\end{equation}

The F1 score is the harmonic mean of precision and recall:

\begin{equation}
    F1~score = 2 \times \frac{\text{Precision} \times \text{Recall}}{\text{Precision} + \text{Recall}},
\end{equation}

The area under the ROC curve measures the ability of the model to distinguish between classes at different thresholds, and can be described by the Riemann–Stieltjes integral for step functions:

\begin{equation}
    AUC\text{--}ROC = \int_{0}^{1} \text{TPR}(x
    ) \, d\text{FPR}(x),
\end{equation}

where TPR (True Positive Rate) and FPR (False Positive Rate) are given by:

\begin{equation}
    \text{TPR} = \frac{TP}{TP + FN}, \quad \text{FPR} = \frac{FP}{FP + TN}.
\end{equation}

These metrics provide an assessment of the classification performance of each model. In the following section, we analyze the results based on these metrics.

\subsection{Ensemble strategy}
\label{sec:ensemble}

Researchers in LSST have already proposed ensembles of five neural networks with predictions from a citizen science project to improve AUC--ROC by testing three methods \citep{Holloway2023ASurveys}: generalized mean, dependent and independent
Bayesian combination.

Our ensemble method aggregates predictions from multiple deep learning architectures during inference using a uniform averaging scheme often referred to as soft voting. Specifically, each model independently outputs a probability score for the presence of a strong lensing event, and we combine these by taking the mean of the predicted probabilities. This strategy reduces variance and can lower bias in individual model predictions if the models are diverse and uncorrelated, improving the detection of subtle lensing features that may be faint or affected by noise and other imaging artifacts.

For \( N = 10 \) total models, the ensemble probability \(p_{\text{ensemble}}\) is computed as the uniform average of individual model probabilities:

\begin{equation}
\label{eq:weightedp}
p_{\text{ensemble}}(x) = \frac{1}{N} \sum_{i=1}^{N} p_i(x),\
\end{equation}

where \(p_i(x)\) represents the predicted probability of the \(i\)-th model for input \(x\). In practice, any model not included (reference ResNet‐18) simply does not contribute to the normalization factor \(N\).

While there are various ensemble methods, including boosting (combining sequentially trained models weighted by their performance), bagging (averaging predictions from models trained on bootstrapped samples), stacking (training a meta-model on predictions from multiple base models), and Bayesian model averaging (weighting models by their posterior probabilities), this study adopts a uniform probability approach as a proof of concept for combining multiple model predictions. More adaptive strategies, such as meta-learning or weighted ensembles based on validation, are beyond the scope of this work.

\section{Results}
\label{sec:results}

\begin{figure*}
    \centering    
    \includegraphics[width=1.0\textwidth]{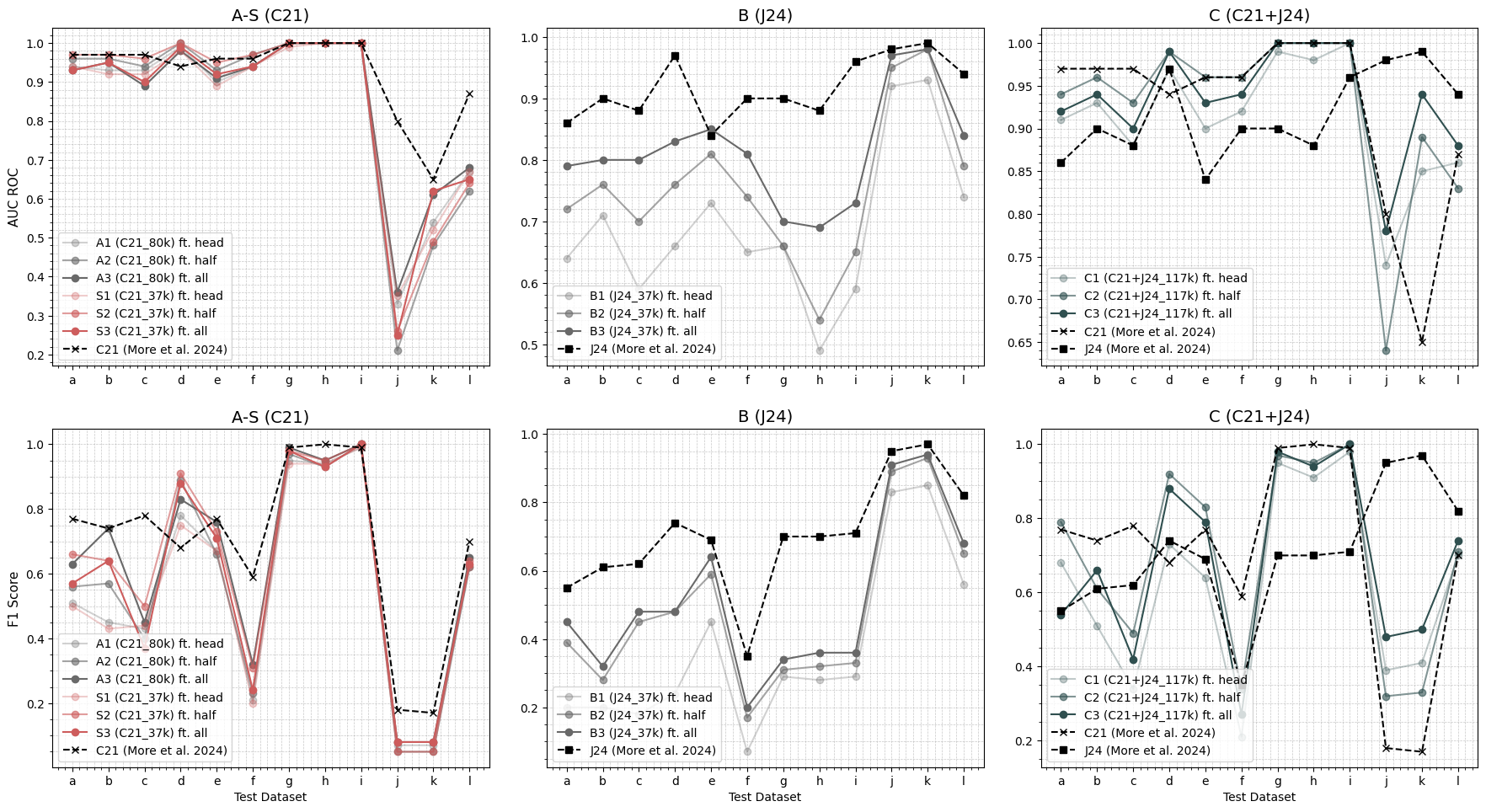}
    \caption{Ensemble results for experiments  $\mathcal{A}$-  $\mathcal{S}$ (C21), $\mathcal{B}$ (J24) and $\mathcal{C}$ (C21+J24). The degree of fine-tuning is depicted by transparency: a lighter hue indicates that only the classifier head is updated, mid-tone color represents retraining half of the architecture, and full opacity means that all layers were unfrozen. Fine-grained tabular results for the ensembles and the references \citep{More2024SystematicLenses} can be found in the Appendix.}
    \label{fig:results}
\end{figure*}

This section presents the classification results of the evaluated models. Figure~\ref{fig:results} provides a graphical summary of the ensemble prediction performance, illustrating AUC--ROC (top) and F1 score (bottom) results across test sets (\textit{a}–\textit{l}) for each training configuration. The plots highlight the effect of fine-tuning strategies (ranging from modifying just the final layer to retraining the full model architecture) as well as the influence of the training dataset. The visual contrast between the A-S (C21), B (J24), and C (C21+J24) experiments emphasizes how the source and composition of the training data influence the generalization performance: models trained on J24 consistently outperform others on L4-based test sets (\textit{j}, \textit{k}), especially in F1 score. These visual results align with the detailed numerical values reported in Tables~\ref{tab:aucroc_integrated} and~\ref{tab:f1_integrated}; see also Appendix~\ref{sec:appendixC} for a complete description.

\subsection{Evaluation}

Table \ref{tab:best_models-all-experiments} summarizes the best-performing models across all test sets (\textit{a}–\textit{l}), reporting AUC--ROC for each. The best result for each test set corresponds to the model, training dataset, and fine-tuning strategy that achieved the highest score. In particular, several test sets (\textit{a}, \textit{b}, \textit{c} and \textit{f}) were best handled using the reduced dataset $\mathcal{S}$ with partial fine-tuning ($\mathcal{S}_2$), highlighting the efficiency of this setup despite using fewer training samples. On the other hand, full fine-tuning on the C21 dataset $\mathcal{A}_3$ proved highly effective, especially for test sets \textit{d}, \textit{g}, \textit{h}, and \textit{i}, where models like ResNet-18 and CaiT reached near-perfect performance (AUC--ROC > 0.999). Test sets \textit{j} and \textit{k} required training on J24 with full fine-tuning ($\mathcal{B}_3$), suggesting that domain-specific adaptation is beneficial in these cases. While CaiT, ResNet-18 and ensemble models yielded strong results across multiple sets, other architectures such as MLP-Mixer, DeiT III and Twins-PCPVT also emerged as top performers. Overall, these results show that a combination of appropriate fine-tuning depth and dataset choice can achieve performance comparable to or exceeding the state of the art, as reported by \citet{More2024SystematicLenses}. Only test set \textit{l}, which aggregates every object in the common test sample, remains slightly below prior benchmarks, indicating it may pose intrinsic challenges for all evaluated models.

The model with the highest AUC--ROC for each test set is listed on Table \ref{tab:best_models-all-experiments}, while the results from all experiments are at Appendix in Tables~\ref{tab:aucroc_integrated}~and~\ref{tab:f1_integrated}. They display the AUC--ROC scores of the ensemble model evaluated on the four different experiments, $\mathcal{A}$, $\mathcal{B}$, $\mathcal{C}$, and $\mathcal{S}$, along with three transfer learning settings for each previously discussed dataset. Moreover, Table~\ref{tab:f1_integrated} presents the F1 scores of the ensemble model evaluated across the same experiments and datasets. To provide further insight into these results, both tables also include the values reported by \citet{More2024SystematicLenses}. A detailed analysis of these tables highlights the strong performance of models trained on the reduced C21 dataset and the effectiveness of partial fine-tuning in several cases. While L3-based test sets are consistently well-classified, test sets involving L4 lenses (\textit{j} and \textit{k}) require training on J24 to achieve competitive results, underscoring the importance of specific features. Additional insights, including comparisons with prior benchmarks and performance variability across datasets, are discussed in the Appendix.

\begin{table}
\centering
\caption{The best model for each test dataset, measuring AUC--ROC. The Experiment column indicates what dataset ($\mathcal{A}$, $\mathcal{B}$, $\mathcal{C}$, or $\mathcal{S}$) was used for training and degree of fine-tuning (1, 2, or 3). Best overall results in bold.}
\label{tab:best_models-all-experiments}
\begin{tabular}{lll|ll}
\hline
\textbf{Set} & \textbf{Exp.} & \textbf{Model} &
\multicolumn{2}{c}{\textbf{AUC--ROC}} \\
\cline{4-5}
& & & Best model & \citet{More2024SystematicLenses} \\
\hline
\textit{a} & $\mathcal{S}_2$ & Ensemble    & $0.97$           & $\mathbf{0.98}$ (S22) \\
\textit{b} & $\mathcal{S}_2$ & Ensemble    & $\mathbf{0.97}$   & $\mathbf{0.97}$ (C21) \\
\textit{c} & $\mathcal{S}_2$ & DeiT~III    & $0.96$           & $\mathbf{0.99}$ (S22) \\
\textit{d} & $\mathcal{A}_3$ & ResNet-18   & $\mathbf{0.\overline{99}}$   & $0.97$ (J24) \\
\textit{e} & $\mathcal{C}_2$ & Ensemble    & $0.95$           & $\mathbf{0.96}$ (C21) \\
\textit{f} & $\mathcal{S}_2$ & Ensemble    & $\mathbf{0.97}$   & $0.96$ (C21, S22) \\
\textit{g} & $\mathcal{A}_3$ & CaiT        & $\mathbf{0.\overline{99}}$ & $\mathbf{1.00}$ (C21) \\
\textit{h} & $\mathcal{A}_2$ & CaiT        & $\mathbf{0.\overline{99}}$ & $\mathbf{1.00}$ (C21, S22) \\
\textit{i} & $\mathcal{A}_3$ & ResNet-18   & $\mathbf{0.\overline{99}}$ & $\mathbf{1.00}$ (C21, S22) \\
\textit{j} & $\mathcal{B}_3$ & Twins-PCPVT & $0.97$           & $\mathbf{0.99}$ (I24) \\
\textit{k} & $\mathcal{B}_3$ & Ensemble    & $0.98$           & $\mathbf{0.99}$ (J24, I24) \\
\textit{l} & $\mathcal{C}_3$ & MLP-Mixer   & $0.92$           & $\mathbf{0.94}$ (J24) \\
\hline
\end{tabular}
\end{table}

\subsection{Inference on L2}

We evaluate trained models in inference mode to classify test subset L2, in order to quantify how many lenses we are able to recover from those identified in HSC PDR2 images of GAMA09H by \citet{More2024SystematicLenses}. For each dataset and fine-tuning configuration, Table \ref{tab:detection_rates} shows the highest detection rate achieved on L2. The MLP-Mixer model reached the best performance (132/138), having a recall of 95.65\%, when trained with C21 and partial fine-tuning ($\mathcal{A}_2$).

\begin{figure*}
    \centering            \includegraphics[width=0.59\textwidth]{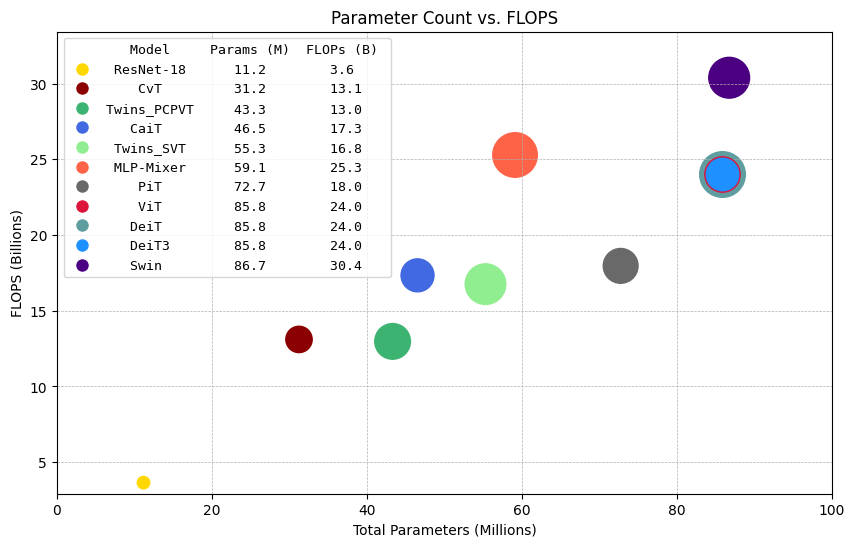}
    \caption{Parameter count vs. floating point operations (FLOPs). Markers are scaled by the mean AUC--ROC, training every layer with sets C21+J24 ($\mathcal{C}_3$). The MLP-Mixer has a good balance between classification performance, in this case AUC--ROC (size of the marker), and its computational complexity (Parameters and FLOPs). Note that ViT, DeiT and DeiT3 models overlap at 85.8 million parameters and 24.0 billion FLOPs.}
    \label{fig:flops}
\end{figure*}

Inference-time computational complexity is assessed using two main metrics: the number of parameters and the number of floating point operations (FLOPs). Networks with more parameters typically have a higher capacity to learn complex patterns; however, they may also be prone to overfitting, failing to generalize to unseen data. FLOPs represent the computational cost of a single forward pass. Both the parameter count and FLOPs are computed using \texttt{ptflops} \citep{ptflops} and are visualized in Figure~\ref{fig:flops}. Note that marker sizes in the figure are scaled according to the mean AUC--ROC.

\begin{table}
\centering
\caption{Inference on Test Subset L2 (138 lenses): Experiment, Model, Detections, and True Positive Rate (TPR). Best performance is marked in bold.}
\label{tab:detection_rates}
\begin{tabular}{l l c c}
\hline
Experiment & Model        & Detections & Recall (\%) \\
\hline
$\mathcal{A}_1$         & Swin         & 126        & 91.30 \\
$\mathcal{A}_2$         & MLP-Mixer    & \textbf{132}        & \textbf{95.65} \\
$\mathcal{A}_3$        & PiT          & 128        & 92.75 \\
\hline
$\mathcal{B}_1$         & ViT          & 69         & 50.00 \\
$\mathcal{B}_2$         & ViT \& DeiT     & 79         & 57.25 \\
$\mathcal{B}_3$         & MLP-Mixer    & 80         & 57.97 \\
\hline
$\mathcal{C}_1$         & Swin         & 120        & 86.96 \\
$\mathcal{C}_2$         & MLP-Mixer    & 128        & 92.75 \\
$\mathcal{C}_3$         & ResNet-18    & 131        & 94.93 \\
\hline
$\mathcal{S}_1$         & Swin         & 124        & 89.86 \\
$\mathcal{S}_2$         & MLP-Mixer \& Ensemble    & 129        & 93.48 \\
$\mathcal{S}_3$         & PiT          & 130        & 94.20 \\
\hline
\end{tabular}
\end{table}

\section{Conclusions}
\label{sec:conclusions}

In this paper we present GraViT, a flexible framework for strong gravitational lens discovery that leverages transfer learning with pre-trained Vision Transformer and MLP-Mixer models. We assess the impact of transfer learning on classification performance by examining data quality, model architecture, training strategies, and ensemble predictions. Our work reproduces experiments in a previous systematic comparison of neural networks \citep{More2024SystematicLenses} and provides insights into the detectability of lenses on this common test sample.

Our experiments with transfer learning demonstrate that unfreezing and retraining deeper layers generally leads to better performance when the downstream task diverges significantly from the ImageNet-21k  \citep{Ridnik2021ImageNet-21KMasses} and ImageNet-1k \citep{Deng2009ImageNet:Database} pretraining. For instance, in this galaxy-galaxy strong gravitational lensing study, models that fine-tune half of the architecture often outperform other settings, achieving a good balance between keeping ImageNet activations and incorporating gravitational lensing features.

With datasets J24 \citep{Jaelani2024SurveyNetworks} and C21 \citep{Canameras2021HOLISMOKESSurvey} ranging from 37,320 -- 80,000 images, our trained models exhibit competitive performance, comparable to the I24 network \citep{ishida24}, trained on the same J24 sample.  Only when combining C21+J24 (117,320 images) do some variants manage to outperform the reported results, as ViTs benefit from large datasets.

Instead, reducing the C21 dataset from \(\text{C21}_{80\,000}\) to \(\text{C21}_{37\,320}\) (the \(\mathcal{S}\) experiment), matching the size of J24, produced unexpected results. Models trained with this smaller sample, as an ensemble (\(\mathcal{S}_2\) in Table~\ref{tab:aucroc_integrated}), achieve similar values for AUC--ROC and F1 score, suggesting noise in data or that features in \(\text{C21}_{80\,000}\) are already present in \(\text{C21}_{37\,320}\), and do not contribute towards generalization.

Neural networks trained with C21 and S22 data excel on test sets \textit{a} to \textit{i}; in contrast, training with J24 tends to achieve higher AUC--ROC and F1 score on \textit{j}, \textit{k} and \textit{l}, an outcome that reflects the experimental design of the common test sample \citep{More2024SystematicLenses, Canameras2021HOLISMOKESSurvey,Shu2022HOLISMOKESProgram, Jaelani2024SurveyNetworks, ishida24}. Therefore, broad generalization of conclusions is a challenge.

The ViT and some of its variants, such as the MLP-Mixer, continue the trend of removing hand-crafted features and inductive biases from models \citep{dosovitskiy2021imageworth16x16words}, depending instead on raw data and large-scale pretraining to achieve state-of-the-art performance.

CNNs rely on strong built-in biases for local feature extraction and translation equivariance, that gradually expand their receptive field to capture complex patterns in a data-efficient manner. In contrast, ViTs use self-attention to establish global context and model long-range dependencies; however, they typically require larger datasets and more computational resources to learn effective spatial relationships.

The MLP-Mixer \citep{Tolstikhin2021MLP-Mixer:Vision} with $\mathcal{C}_3$ setting achieved the best results on test set \textit{l}, which aggregates all of the samples, and recovered the most lenses (95.65\%) doing inference on L2 with $\mathcal{A}_2$. Its computational complexity is linear in the number of input patches $\mathcal{O}(n)$, as opposed to the quadratic $\mathcal{O}(n^2)$ self-attention mechanism. As the token-mixing MLP scales linearly with the number of patches, and patch size is treated as a fixed hyperparameter, the overall computational cost grows linearly with the number of pixels. This allows the model to efficiently scale to high-resolution astronomical data.

Ensemble strategies address the bias-variance trade-off by combining predictions from multiple models. The ensemble reduces variance, and if these models are diverse and independent, it counters individual model biases by mitigating systematic errors.

This evaluation demonstrates the impact of transfer learning by fine-tuning on domain-specific image data that exhibits the unique features of galaxy-galaxy strong gravitational lensing. While developed for lens detection, this framework can be adapted to other domains and classification tasks, such as the identification of supernovae via triplets from difference image analysis (DIA). Our pipeline, based on \texttt{PyTorch}, \texttt{astropy}, and \texttt{timm}, provides an efficient transfer learning solution for wide-field astronomical surveys by reusing compute from large-scale pre-trained models and exploiting fine-tuning with selective unfreezing of the neural architecture.

\section*{Acknowledgements}

R.P. thanks Secretaría de Ciencia, Humanidades, Tecnología e Innovación (SECIHTI) for financial support under grant no.\ 776448, and Universidad Autónoma de San Luis Potosí (UASLP) for supporting research in machine learning for astrophysics, and our collaboration with Instituto de Astronomía, Universidad Nacional Autónoma de México (IA-UNAM), LSST-MX consortium, and the Legacy Survey of Space and Time (LSST) at the Vera C. Rubin Observatory.

The authors are grateful to the IDAC at the IA-UNAM for providing high-performance computing resources for this model evaluation. Some of the calculations in this work were carried out on the HPC clusters Atocatl and Tochtli at LAMOD-UNAM. LAMOD is a collaborative project of IA, ICN, IQ Institutes and DGTIC at UNAM.

A.T.J. is supported by the Program Penelitian, Pengabdian Masyarakat, dan Inovasi (PPMI) ITB 2025.

R.P. gratefully acknowledges Dr. Raoul Cañameras for providing the dataset from HOLISMOKES VI to evaluate neural architectures.

O.V., J.C.C. and R.P. acknowledge support from the SECIHTI grant CF-2023-G-1052 and DGAPA UNAM grants  AG102123 and IG101725.


\section*{Data Availability}

The datasets are available upon reasonable request to Dr. Anupreeta More. The open-source code can be found on GitHub\footnote{\url{https://github.com/parlange/gravit}}, and the models are publicly available on  Hugging Face\footnote{\url{https://huggingface.co/parlange}} and Zenodo\footnote{\url{https://zenodo.org/}}.




\bibliographystyle{mnras}
\bibliography{references}



\appendix

\section{Common Test Sample}\label{sec:appendixA}

\begin{figure*}
    \centering
    \begin{minipage}{0.49\textwidth}
        \centering
        \includegraphics[width=\textwidth]{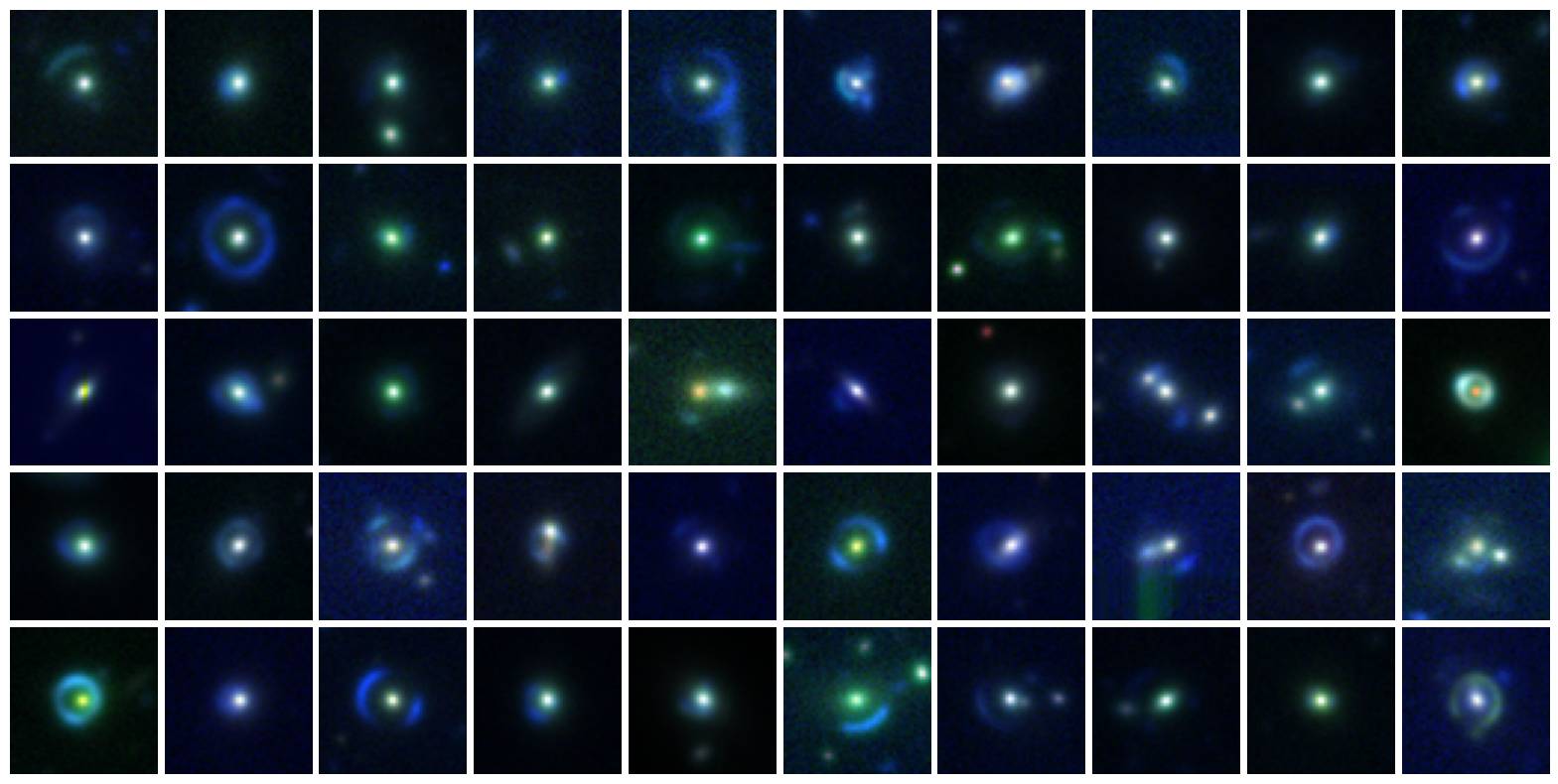}
        \caption{Common test sample: lenses in \textit{i} \citep{More2024SystematicLenses}.}
        \label{fig:i-lens}
    \end{minipage}
    \hfill
    \begin{minipage}{0.49\textwidth}
        \centering
        \includegraphics[width=\textwidth]{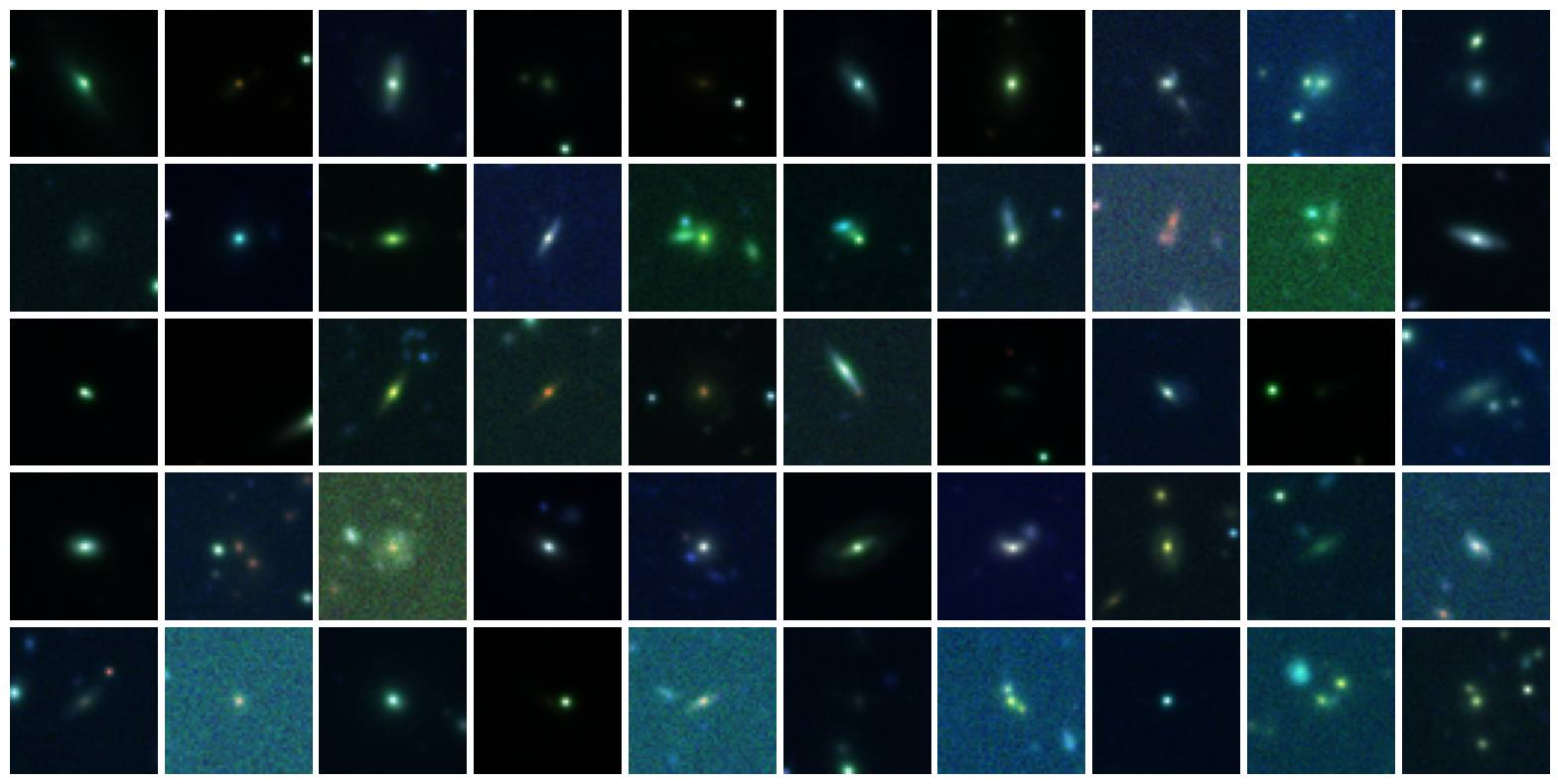}
        \caption{Common test sample: non-lenses in \textit{i} \citep{More2024SystematicLenses}.}
        \label{fig:i-nonlens}
    \end{minipage}
    \label{fig:i-lens-nonlens}
\end{figure*}

\begin{figure*}
    \centering
    \begin{minipage}{0.49\textwidth}
        \centering
        \includegraphics[width=\textwidth]{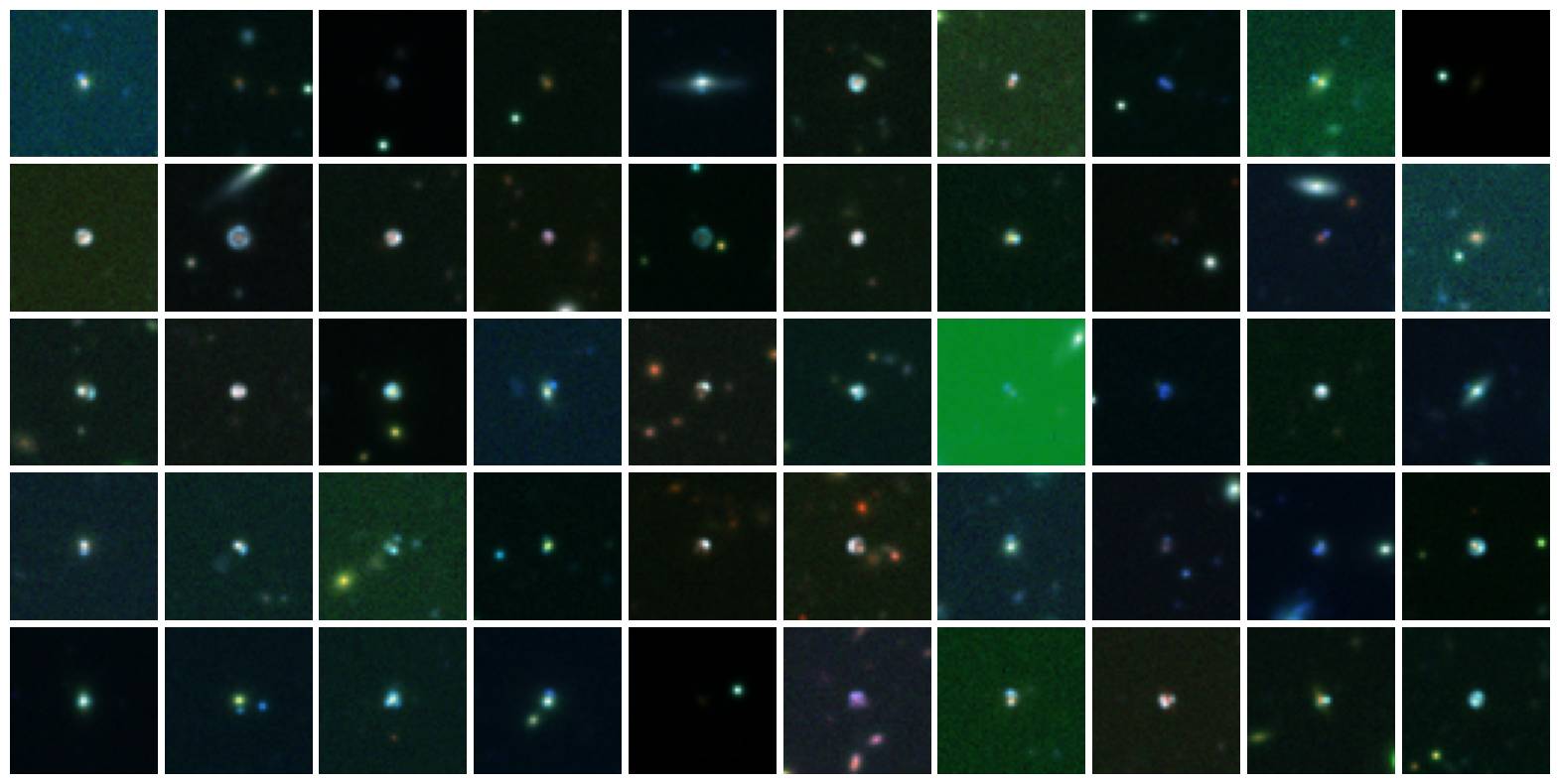}
        \caption{Common test sample: lenses in \textit{j} \citep{More2024SystematicLenses}.}
        \label{fig:j-lens}
    \end{minipage}
    \hfill
    \begin{minipage}{0.49\textwidth}
        \centering
        \includegraphics[width=\textwidth]{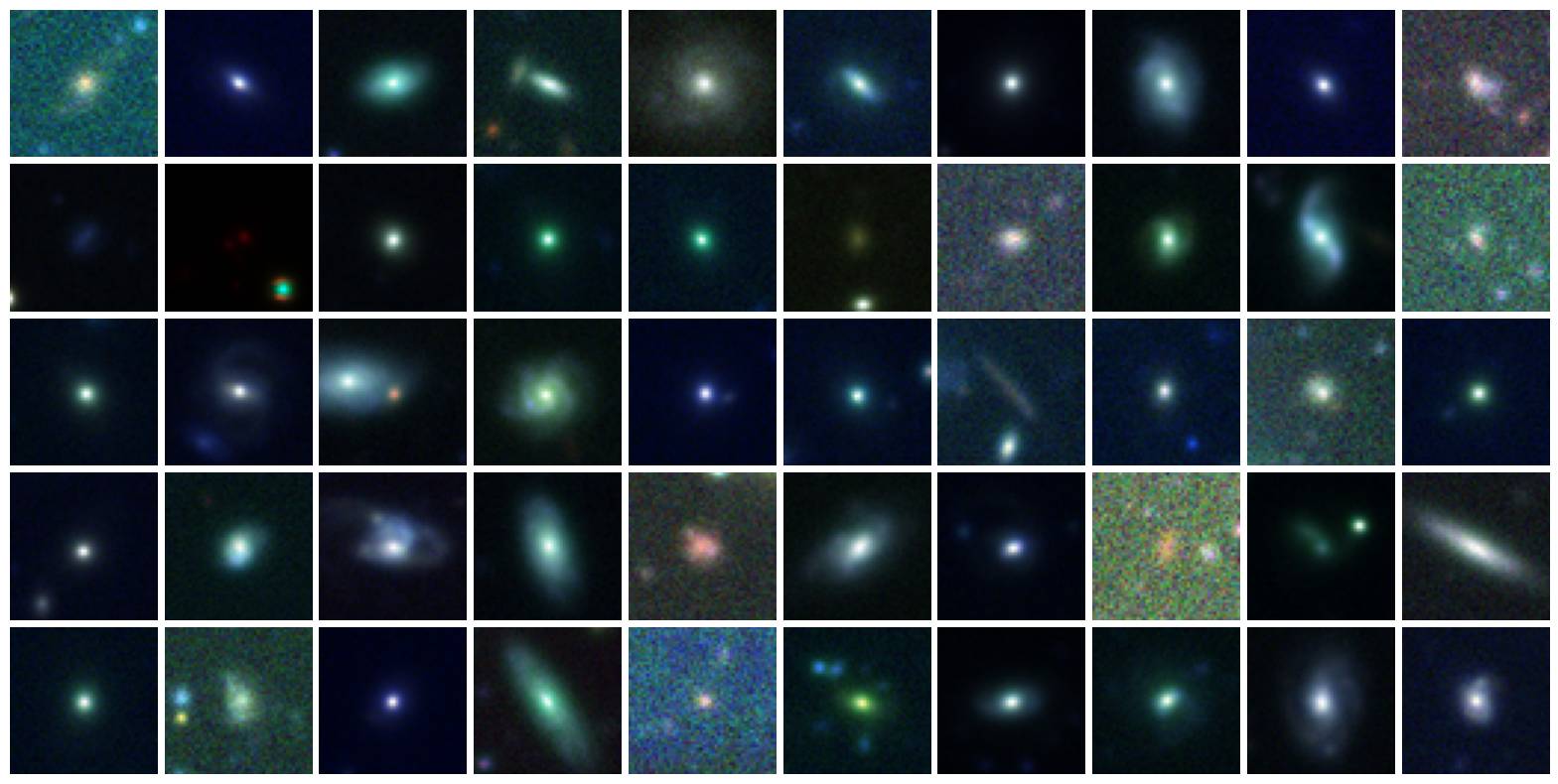}
        \caption{Common test sample: non-lenses in \textit{j} \citep{More2024SystematicLenses}.}
        \label{fig:j-nonlens}
    \end{minipage}
    \label{fig:j-lens-nonlens}
\end{figure*}

\begin{figure*}
    \centering
    \begin{minipage}{0.49\textwidth}
        \centering
        \includegraphics[width=\textwidth]{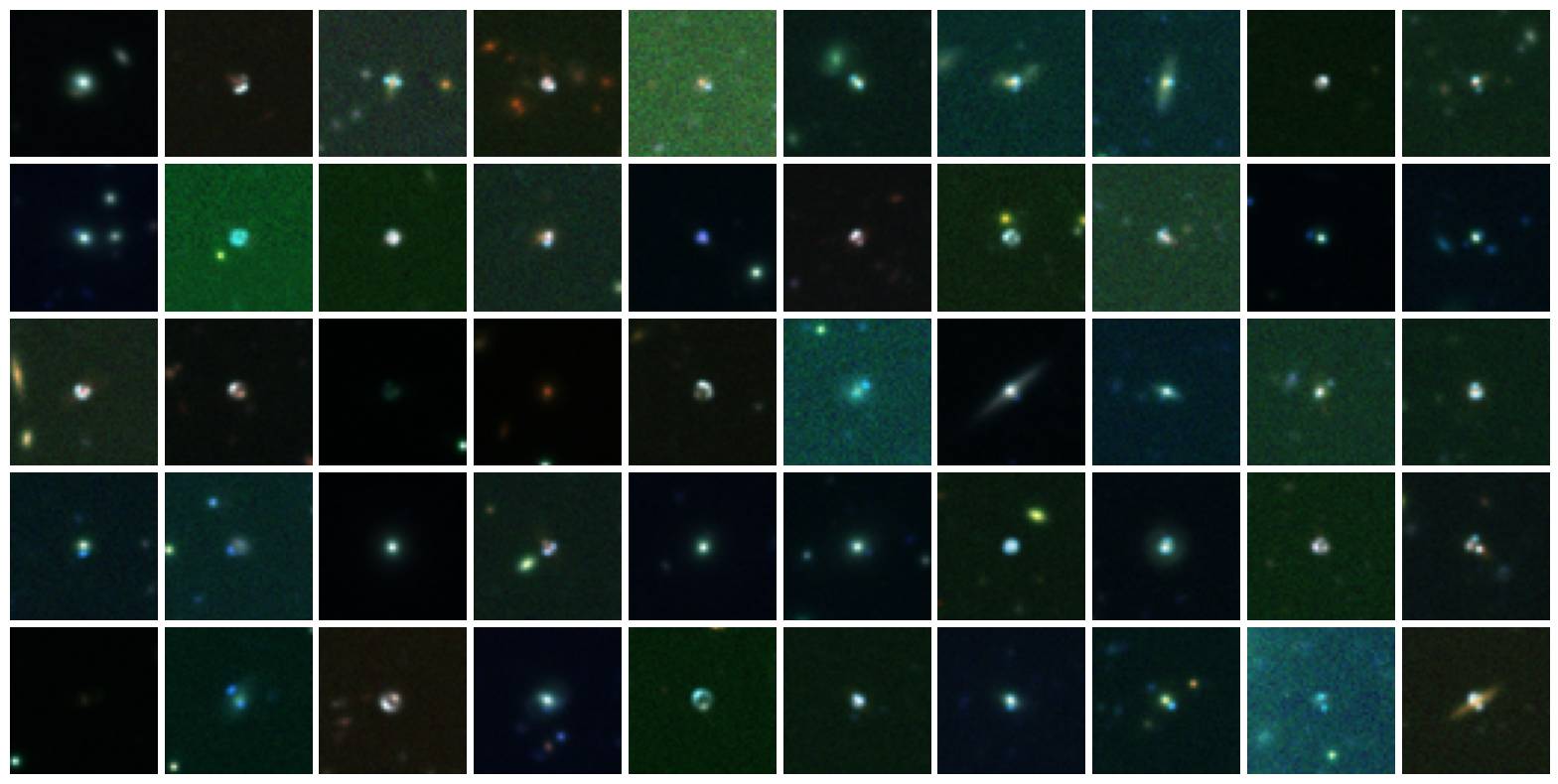}
        \caption{Common test sample: lenses in \textit{k} \citep{More2024SystematicLenses}.}
        \label{fig:k-lens}
    \end{minipage}
    \hfill
    \begin{minipage}{0.49\textwidth}
        \centering
        \includegraphics[width=\textwidth]{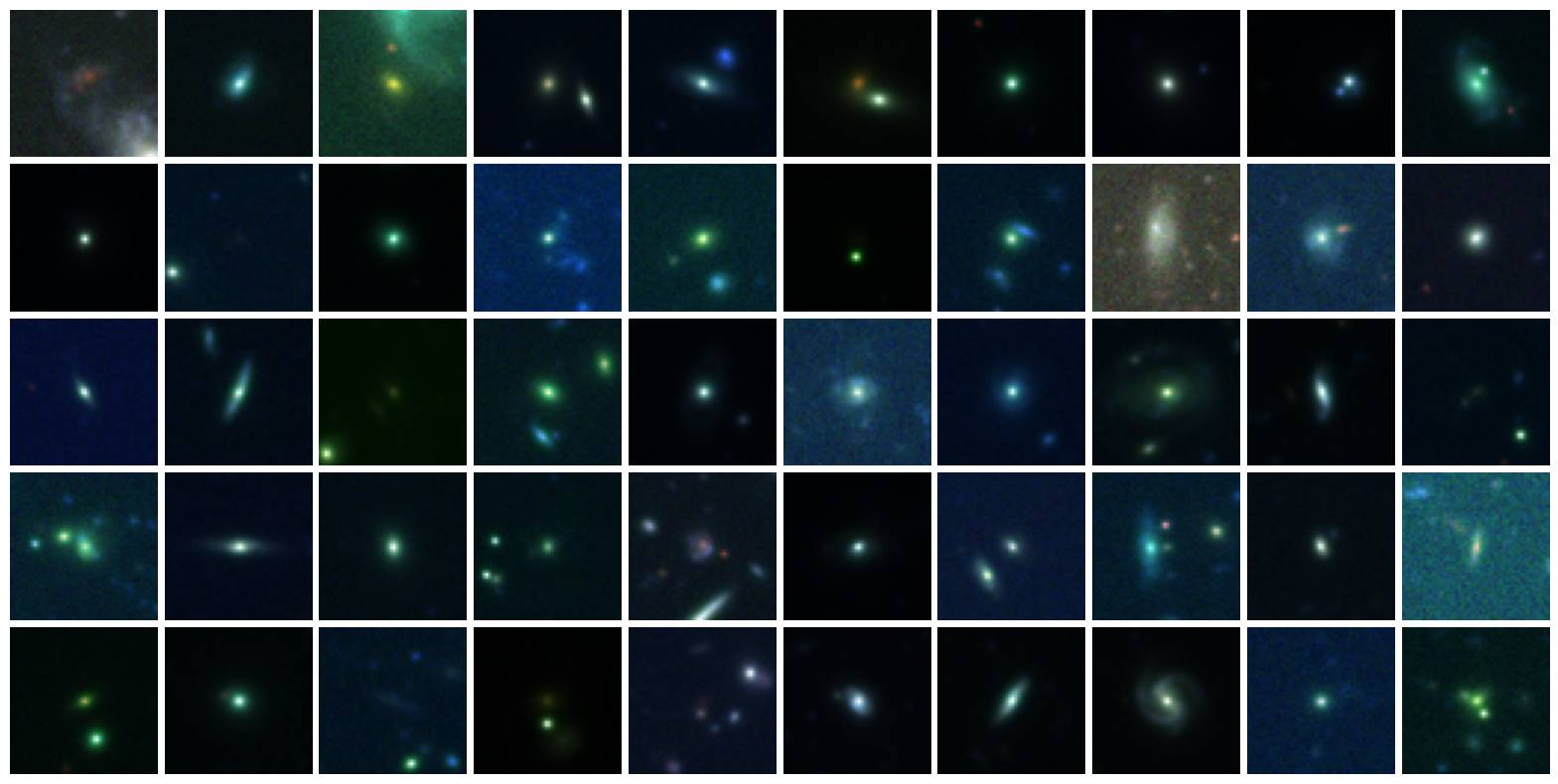}
        \caption{Common test sample: non-lenses in \textit{k} \citep{More2024SystematicLenses}.}
        \label{fig:k-nonlens}
    \end{minipage}
    \label{fig:k-lens-nonlens}
\end{figure*}

\begin{figure*}
    \centering
    \begin{minipage}{0.49\textwidth}
        \centering
        \includegraphics[width=\textwidth]{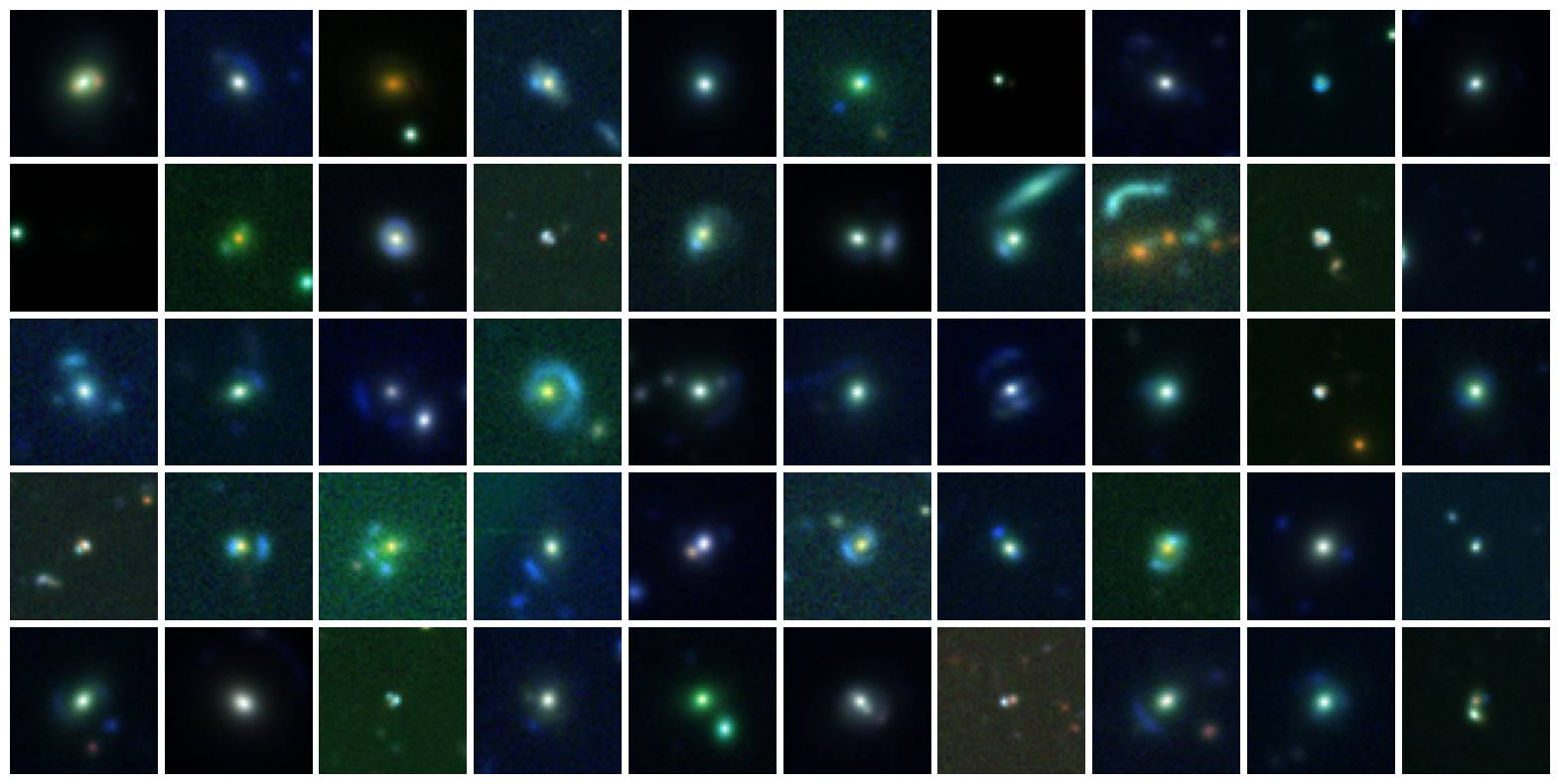}
        \caption{Common test sample: lenses in \textit{l} \citep{More2024SystematicLenses}.}
        \label{fig:l-lens}
    \end{minipage}
    \hfill
    \begin{minipage}{0.49\textwidth}
        \centering
        \includegraphics[width=\textwidth]{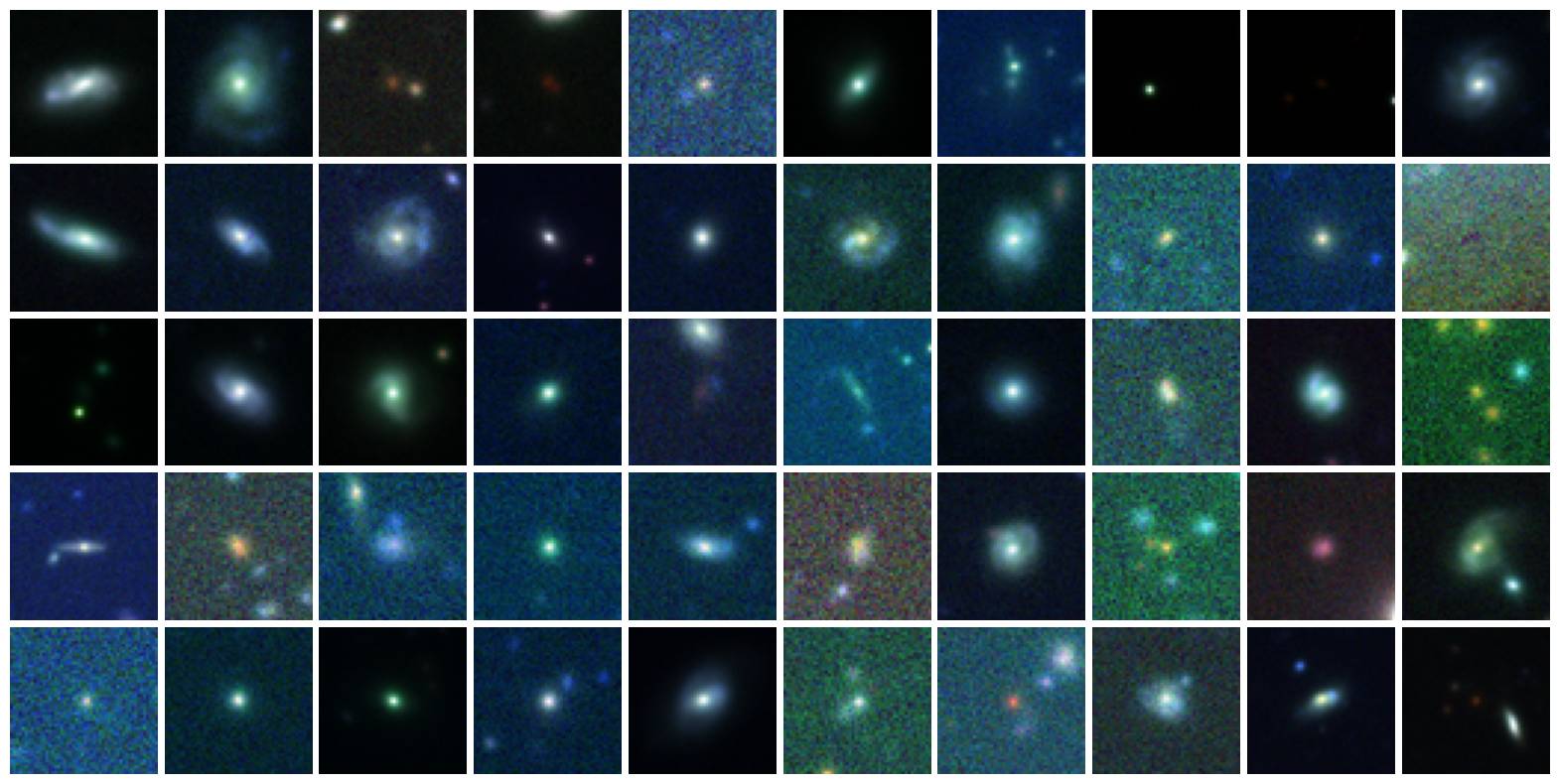}
        \caption{Common test sample: non-lenses in \textit{l} \citep{More2024SystematicLenses}.}
        \label{fig:l-nonlens}
    \end{minipage}
    \label{fig:l-lens-nonlens}
\end{figure*}

A representative, random sample of fifty lenses (Figures \ref{fig:i-lens}, \ref{fig:j-lens}, \ref{fig:k-lens} and \ref{fig:l-lens}) and fifty non-lenses (Figures \ref{fig:i-nonlens}, \ref{fig:j-nonlens}, \ref{fig:k-nonlens} and \ref{fig:l-nonlens}) is shown for visual inspection of test datasets \textit{i}, the easiest, and the more challenging \textit{j}, \textit{k}, and aggregated test set \textit{l} \citep{More2024SystematicLenses}.

\section{Data Provenance}\label{sec:appendixB}

Table~\ref{tab:comparison} summarizes the main characteristics of the HOLISMOKES VI \citep{Canameras2021HOLISMOKESSurvey} and SuGOHI X \citep{Jaelani2024SurveyNetworks} lens searches. Both surveys operate on the same HSC-SSP PDR2 Wide footprint ($\sim800\,\mathrm{deg}^2$, \textit{gri}), but adopt fundamentally different strategies. HOLISMOKES VI performs a broad search over 62.5 million galaxies using only minimal quality cuts (i-band Kron radius $\ge0.8''$, basic artifact flagging), generates 40,000 GLEE-based simulated SIE+shear lenses, and trains a ResNet-18 variant to select candidates at $p_{\rm ResNet}>0.1$, achieving a false-positive rate $\lesssim0.01\%$ and recovering 206 new high-confidence systems (plus 88 upgraded) and 173 known lenses. In contrast, SuGOHI X applies stringent preselection on photometric redshift ($0.2<z_{\rm phot}<1.2$), stellar mass ($M_*>5\times10^{10}\,M_\odot$), low specific star formation rate, and morphology flags to reduce its parent sample to 2.35 million galaxies, uses 18,660 \texttt{SIMCT}-based simulated lenses and a custom 5-layer CNN with a high threshold ($p_{\rm CNN}>0.9$), yielding 43 Grade A, 269 Grade B, and 880 Grade C discoveries (504 known) with an AUC--ROC $\approx0.976$.

\begin{table*}
\centering
\caption{Comparison of gravitational lens search methodologies in HOLISMOKES VI \citep{Canameras2021HOLISMOKESSurvey} and SuGOHI X \citep{Jaelani2024SurveyNetworks}. Both works use simulated lens systems overlaid on real HSC-SSP data, but differ in scope, architecture, and selection strategy. HOLISMOKES VI avoids strict preselection and conducts a broad search across 62.5 million galaxies, using only minimal quality cuts such as i-band Kron radius $\geq$ 0.8$''$ and artifact flagging, aiming to maximize completeness. In contrast, SuGOHI X applies stringent preselection based on photometric redshift ($0.2 < z_\mathrm{phot} < 1.2$), stellar mass ($M_* > 5 \times 10^{10}\, M_\odot$), low specific star formation rate, morphology flags, and additional quality filters, reducing the search sample to 2.35 million galaxies.}
\label{tab:comparison}
\begin{tabular}{lcc}
\hline
\textbf{Aspect} & \textbf{HOLISMOKES VI} & \textbf{SuGOHI X} \\
\hline
Identifier & C21 ($\mathcal{A}$) & J24 ($\mathcal{B}$) \\

Survey & HSC-SSP PDR2 Wide (800 deg$^2$, \textit{gri}) & HSC-SSP PDR2 Wide (800 deg$^2$, \textit{gri}) \\
Number of galaxies searched & 62.5 million & 2.35 million \\
Image cutout size & $12'' \times 12''$ & $64 \times 64$ pixels ($\sim$10.8$''$) \\
Image source & Real HSC \textit{gri} cutouts (PDR2) & Real HSC \textit{gri} cutouts (PDR2) \\
Simulated lenses & 40,000 \texttt{GLEE}-based lenses (SIE + shear) & 18,660 \texttt{SIMCT}-based lenses (SIE + shear) \\
Background source models & Galaxies from Hubble UDF (spectroscopic $z$) & Sérsic profiles from CFHTLenS colors \\
Lens model parameters & SIE with $0.75'' < \theta_E < 2.5''$, $0.4 < z_\ell < 0.7$ & SIE with $0.5'' < \theta_E < 3.0''$, $0.2 < z_\ell < 1.1$ \\

Einstein radius distribution & Uniform prior in $0.75''$–$2.5''$ & SIMCT-derived distribution (accepted if $0.5''$–$3.0''$) \\

Non-lens examples & Spirals, LRGs, compact groups & Random galaxies, spirals, stars, groups \\
CNN architecture & ResNet-18 variant (deep residual network) & 5-layer custom CNN (6.4M parameters) \\
Training set size & 40,000 lenses + 40,000 non-lenses & 14,928 lenses + 14,928 non-lenses \\
Validation set & 500 lenses + 500 non-lenses & 3,732 lenses + 3,732 non-lenses \\
Test set & 202 SuGOHI lenses + 91,000 COSMOS non-lenses & 220 SuGOHI lenses + 78,665 non-lenses \\
Detection threshold & $p_\mathrm{ResNet} > 0.1$ for initial candidate selection & $p_\mathrm{CNN} > 0.9$ \\
Final candidates & 206 new high-confidence + 88 upgraded & 43 grade A, 269 grade B, 880 grade C \\
Overlap with known lenses & 173 known systems recovered & 504 known systems recovered \\
Key feature & Large-scale, generalizable search with low FPR & High precision (AUC = 0.976), careful preselection \\
\hline
\end{tabular}
\end{table*}

\section{AUC--ROC and F1 score Tables for Ensembles in All Experiments}\label{sec:appendixC}

Table~\ref{tab:aucroc_integrated} presents AUC--ROC scores for the ensemble model evaluated on each test dataset and transfer learning setting. Partial fine-tuning on the reduced C21 dataset ($\mathcal{S}_2$) yields excellent performance, matching or surpassing benchmarks for several test sets (\textit{a}–\textit{f}). Test sets involving L3 (\textit{g}–\textit{i}) are consistently well-handled by most models, with AUC--ROC scores reaching 1.00 across multiple experiments. In contrast, test sets \textit{j} and \textit{k} (defined by L4 lenses) show that only models trained on J24 ($\mathcal{B}_3$) perform strongly, achieving 0.97 and 0.98, respectively, while other training sets underperformed significantly. These findings emphasize the impact of both the training dataset and the fine-tuning strategy on generalization performance. The integrated test set (\textit{l}) remains challenging, with ensemble models falling slightly short of the 0.94 reference result set by J24.

Table~\ref{tab:f1_integrated} shows the F1 scores of the ensemble model evaluated on each test dataset and transfer learning settings. In contrast to the AUC--ROC results, F1 scores provide a more rigorous view of performance, particularly in the presence of class imbalance. Test sets based on L3 lenses (\textit{g} – \textit{i}) consistently yield high F1, with scores exceeding 0.95 across multiple configurations. The combination of datasets ($\mathcal{C}$) with partial fine-tuning ($\mathcal{C}_2$) proves particularly effective, leading to top scores in several sets (\textit{a}, \textit{d} and \textit{e}). In contrast, test sets \textit{j} and \textit{k} (involving L4 lenses) show a dramatic performance drop for models trained on C21 or its variants. Only models trained on J24 ($\mathcal{B}_2$, $\mathcal{B}_3$) achieve competitive F1 scores in those cases, highlighting a strong dependence on data provenance. Interestingly, test set \textit{f} (which includes all non-lens classes) shows low F1 scores across the board, suggesting challenges in generalizing to more diverse negative samples. Lastly, performance on the fully-integrated test set (\textit{l}) remains moderate, with the best configuration ($\mathcal{C}_3$) yielding 0.74, slightly ahead of previous benchmarks.

\begin{table*}
\centering
\caption{AUC--ROC scores of the ensemble model evaluated on each test dataset (\textit{a} - \textit{l}) and transfer learning experiment. Values reported by \citet{More2024SystematicLenses} are marked by $^{\dagger}$. Experiments are abbreviated as follows: $\mathcal{A}$ ~C21, $\mathcal{B}$ ~J24, $\mathcal{C}$~C21+J24. An additional experiment $\mathcal{S}$ reduces the sample size of C21 from 80,000 to 37,320. Transfer learning experiments involve fine-tuning: (1)~classification head only, (2)~half of the architecture, and (3)~the entire architecture.}
\begin{tabular}{l c c c c c c c c c c c c | c c c c}
\hline
\textbf{Test Dataset} & $\mathcal{A}_1$ & $\mathcal{A}_2$ & $\mathcal{A}_3$ & $\mathcal{B}_1$ & $\mathcal{B}_2$ & $\mathcal{B}_3$ & $\mathcal{C}_1$ & $\mathcal{C}_2$ & $\mathcal{C}_3$ & $\mathcal{S}_1$ & $\mathcal{S}_2$ & $\mathcal{S}_3$ & C21$^{\dagger}$ & S22$^{\dagger}$ & J24$^{\dagger}$ & I24$^{\dagger}$ \\
\hline
\textit{a}) L1 + L2 - N1 & 0.94 & 0.96 & 0.93 & 0.64 & 0.72 & 0.79 & 0.91 & 0.94 & 0.92 & 0.94 & \textbf{0.97} & 0.93 & 0.97 & 0.98 & 0.86 & 0.80 \\
\textit{b}) L1 + L2 - N2 & 0.93 & 0.96 & 0.95 & 0.71 & 0.76 & 0.80 & 0.93 & 0.96 & 0.94 & 0.92 & \textbf{0.97} & 0.95 & \textbf{0.97} & 0.91 & 0.90 & 0.80 \\
\textit{c}) L1 + L2 - N3 & 0.93 & 0.94 & 0.89 & 0.59 & 0.70 & 0.80 & 0.88 & 0.93 & 0.90 & 0.92 & 0.96 & 0.90 & 0.97 & \textbf{0.99} & 0.88 & 0.83 \\
\textit{d}) L1 + L2 - N4 & 0.99 & \textbf{1.00} & 0.98 & 0.66 & 0.76 & 0.83 & 0.97 & 0.99 & 0.99 & 0.98 & \textbf{1.00} & 0.99 & 0.94 & 0.96 & \textbf{0.97} & 0.87 \\
\textit{e}) L1 + L2 - N5 & 0.90 & 0.93 & 0.91 & 0.73 & 0.81 & 0.85 & 0.90 & \textbf{0.96} & 0.93 & 0.89 & 0.95 & 0.92 & \textbf{0.96} & 0.90 & 0.84 & 0.79 \\
\textit{f}) L1 + L2 - N(all) & 0.94 & \textbf{0.97} & 0.94 & 0.65 & 0.74 & 0.81 & 0.92 & 0.96 & 0.94 & 0.94 & \textbf{0.97} & 0.94 & 0.96 & 0.96 & 0.90 & 0.82 \\
\textit{g}) L3 - N2  & \textbf{1.00} & \textbf{1.00} & \textbf{1.00} & 0.66 & 0.66 & 0.70 & 0.99 & \textbf{1.00} & \textbf{1.00} & 0.99 & \textbf{1.00} & \textbf{1.00} & \textbf{1.00} & 0.99 & 0.90 & 0.71 \\
\textit{h}) L3 - N3  & \textbf{1.00} & \textbf{1.00} & \textbf{1.00} & 0.49 & 0.54 & 0.69 & 0.98 & \textbf{1.00} & \textbf{1.00} & \textbf{1.00} & \textbf{1.00} & \textbf{1.00} & \textbf{1.00} & \textbf{1.00} & 0.88 & 0.76 \\
\textit{i}) L3 - N4  & \textbf{1.00} & \textbf{1.00} & \textbf{1.00} & 0.59 & 0.65 & 0.73 & \textbf{1.00} & \textbf{1.00} & \textbf{1.00} & \textbf{1.00} & \textbf{1.00} & \textbf{1.00} & \textbf{1.00} & \textbf{1.00} & 0.96 & 0.82 \\
\textit{j}) L4 - N2  & 0.33 & 0.21 & 0.36 & 0.92 & 0.95 & 0.97 & 0.74 & 0.64 & 0.78 & 0.35 & 0.26 & 0.25 & 0.80 & 0.70 & 0.98 & \textbf{0.99} \\
\textit{k}) L4 - N4  & 0.54 & 0.48 & 0.61 & 0.93 & 0.98 & 0.98 & 0.85 & 0.89 & 0.94 & 0.52 & 0.49 & 0.62 & 0.65 & 0.81 & \textbf{0.99} & \textbf{0.99} \\
\textit{l}) L (all) - N (all) & 0.67 & 0.62 & 0.68 & 0.74 & 0.79 & 0.84 & 0.86 & 0.83 & 0.88 & 0.67 & 0.64 & 0.65 & 0.87 & 0.91 & \textbf{0.94} & 0.87 \\
\hline
\end{tabular}
\label{tab:aucroc_integrated}
\end{table*}

\begin{table*}
\centering
\caption{F1 scores of the ensemble model evaluated on each test dataset (\textit{a} -- \textit{l}) and transfer learning experiment. Values reported by \citet{More2024SystematicLenses} are marked by $^{\dagger}$. Experiments are abbreviated as follows: $\mathcal{A}$~C21, $\mathcal{B}$~J24, $\mathcal{C}$~C21+J24. An additional experiment $\mathcal{S}$ reduces the sample size of C21 from 80,000 to 37,320. Transfer learning experiments involve fine-tuning: (1)~classification head only, (2)~half of the architecture, and (3)~the entire architecture.}
\begin{tabular}{l c c c c c c c c c c c c | c c c c}
\hline
\textbf{Test Dataset} & $\mathcal{A}_1$ & $\mathcal{A}_2$ & $\mathcal{A}_3$ & $\mathcal{B}_1$ & $\mathcal{B}_2$ & $\mathcal{B}_3$ & $\mathcal{C}_1$ & $\mathcal{C}_2$ & $\mathcal{C}_3$ & $\mathcal{S}_1$ & $\mathcal{S}_2$ & $\mathcal{S}_3$ & C21$^{\dagger}$ & S22$^{\dagger}$ & J24$^{\dagger}$ & I24$^{\dagger}$ \\
\hline
\textit{a}) L1 + L2 – N1 & 0.51 & 0.56 & 0.63 & 0.20 & 0.39 & 0.45 & 0.68 & \textbf{0.79} & 0.54 & 0.50 & 0.66 & 0.57 & 0.77 & 0.76 & 0.55 & 0.45 \\
\textit{b}) L1 + L2 – N2 & 0.45 & 0.57 & \textbf{0.74} & 0.20 & 0.28 & 0.32 & 0.51 & 0.61 & 0.66 & 0.43 & 0.64 & 0.64 & \textbf{0.74} & 0.53 & 0.61 & 0.47 \\
\textit{c}) L1 + L2 – N3 & 0.43 & 0.40 & 0.45 & 0.18 & 0.45 & 0.48 & 0.34 & 0.49 & 0.42 & 0.44 & 0.50 & 0.37 & \textbf{0.78} & 0.77 & 0.62 & 0.48 \\
\textit{d}) L1 + L2 – N4 & 0.78 & 0.89 & 0.83 & 0.24 & 0.48 & 0.48 & 0.73 & \textbf{0.92} & 0.88 & 0.75 & 0.91 & 0.88 & 0.68 & 0.72 & 0.74 & 0.51 \\
\textit{e}) L1 + L2 – N5 & 0.67 & 0.66 & 0.76 & 0.45 & 0.59 & 0.64 & 0.64 & \textbf{0.83} & 0.79 & 0.67 & 0.73 & 0.71 & 0.77 & 0.70 & 0.69 & 0.52 \\
\textit{f}) L1 + L2 – N(all) & 0.21 & 0.23 & 0.32 & 0.07 & 0.17 & 0.20 & 0.21 & 0.34 & 0.27 & 0.20 & 0.31 & 0.24 & \textbf{0.59} & 0.47 & 0.35 & 0.37 \\
\textit{g}) L3 – N2 & 0.95 & 0.97 & \textbf{0.99} & 0.29 & 0.31 & 0.34 & 0.95 & 0.97 & 0.98 & 0.94 & 0.98 & 0.98 & \textbf{0.99} & 0.94 & 0.70 & 0.25 \\
\textit{h}) L3 – N3 & 0.94 & 0.93 & 0.95 & 0.28 & 0.32 & 0.36 & 0.91 & 0.95 & 0.94 & 0.94 & 0.95 & 0.93 & \textbf{1.00} & 0.96 & 0.70 & 0.25 \\
\textit{i}) L3 – N4 & 0.99 & \textbf{1.00} & \textbf{1.00} & 0.29 & 0.33 & 0.36 & 0.98 & \textbf{1.00} & \textbf{1.00} & 0.99 & \textbf{1.00} & \textbf{1.00} & 0.99 & 0.96 & 0.71 & 0.25 \\
\textit{j}) L4 – N2 & 0.07 & 0.05 & 0.08 & 0.83 & 0.89 & 0.91 & 0.39 & 0.32 & 0.48 & 0.08 & 0.05 & 0.08 & 0.18 & 0.30 & \textbf{0.95} & \textbf{0.95} \\
\textit{k}) L4 – N4 & 0.07 & 0.05 & 0.08 & 0.85 & 0.93 & 0.94 & 0.41 & 0.33 & 0.50 & 0.08 & 0.05 & 0.08 & 0.17 & 0.31 & \textbf{0.97} & 0.96 \\
\textit{l}) L(all) – N(all) & 0.62 & 0.62 & 0.65 & 0.56 & 0.65 & 0.68 & 0.70 & 0.71 & 0.74 & 0.62 & 0.64 & 0.63 & 0.70 & 0.70 & \textbf{0.82} & 0.68 \\
\hline
\end{tabular}
\label{tab:f1_integrated}
\end{table*}

\section{ROC Curves}\label{sec:appendixD}

\begin{figure}
    \centering
    \begin{minipage}{0.49\textwidth}
        \centering
        \includegraphics[width=\textwidth]{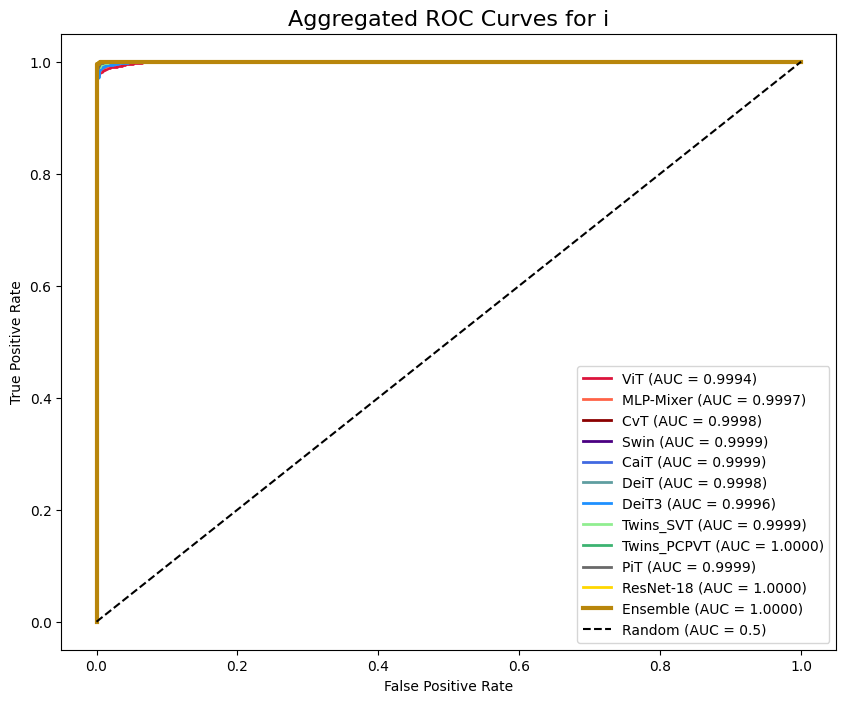}
        \caption{ROC curves for \textit{i} test dataset in $\mathcal{C}_3$ experiment (C21+J24).}
        \label{fig:aggregated_roc_i}
    \end{minipage}
    \hfill
    \begin{minipage}{0.49\textwidth}
        \centering
        \includegraphics[width=\textwidth]{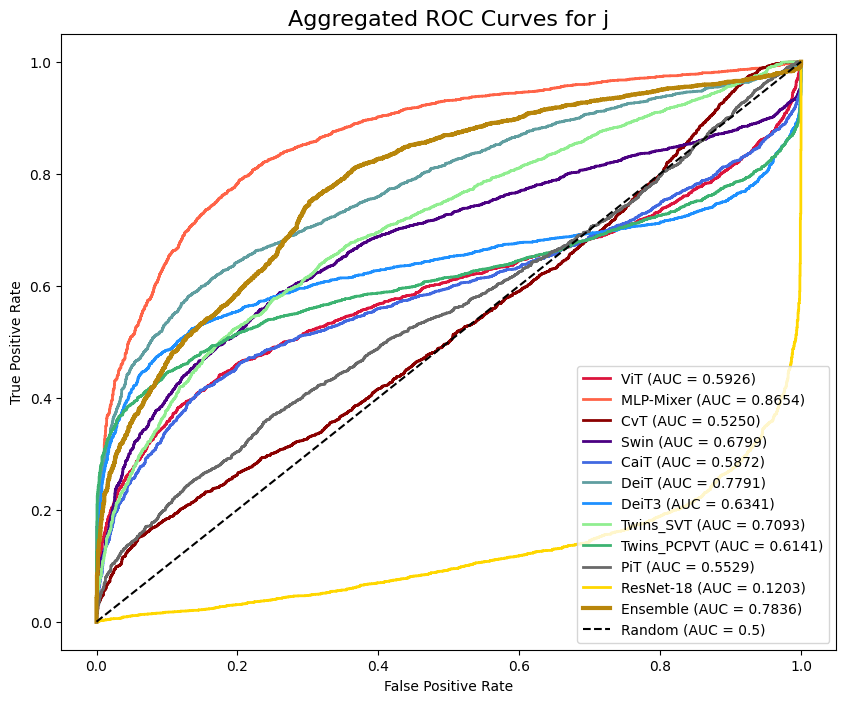}
        \caption{ROC curves for \textit{j} test dataset in experiment $\mathcal{C}_3$ (C21+J24).}
        \label{fig:aggregated_roc_j}
    \end{minipage}
    \label{fig:aggregated_roc_comparison_ij}
\end{figure}

\begin{figure}
    \centering
    \begin{minipage}{0.49\textwidth}
        \centering
        \includegraphics[width=\textwidth]{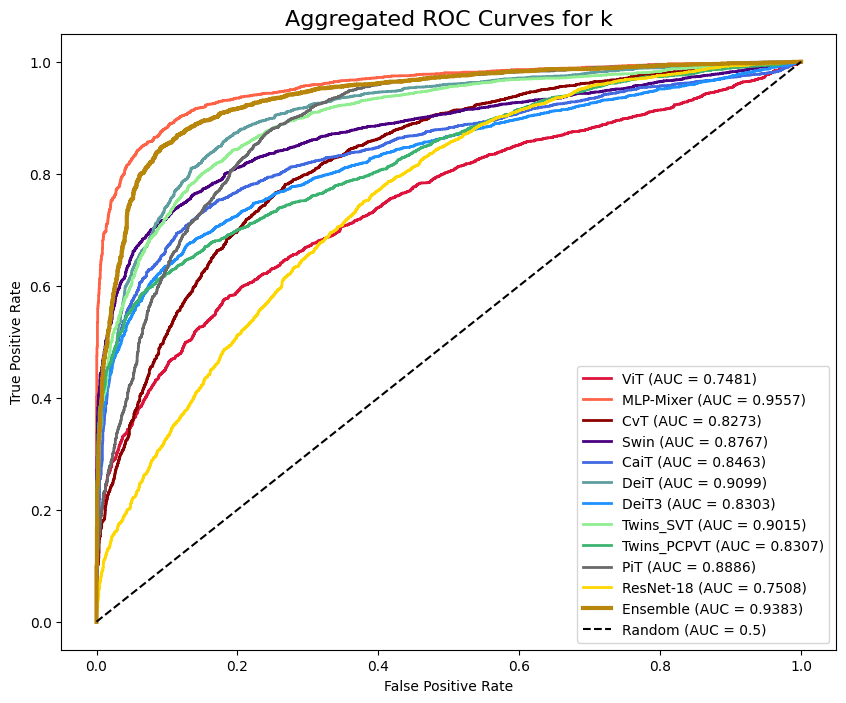}
        \caption{ROC curves for \textit{k} test dataset in $\mathcal{C}_3$ experiment (C21+J24).}
        \label{fig:aggregated_roc_k}
    \end{minipage}
    \hfill
    \begin{minipage}{0.49\textwidth}
        \centering
        \includegraphics[width=\textwidth]{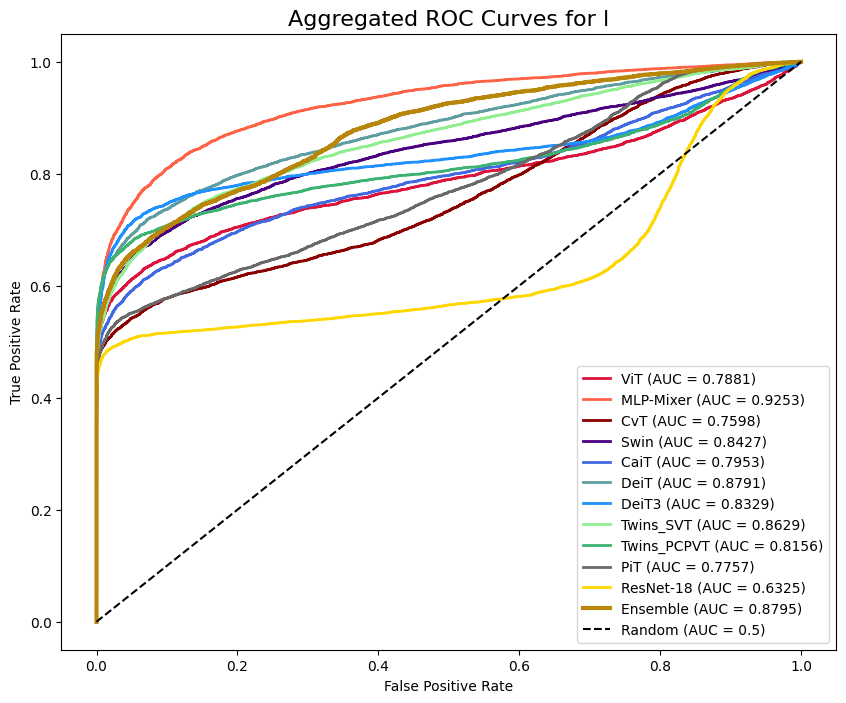}
        \caption{ROC curves for \textit{l} test dataset in experiment $\mathcal{C}_3$ (C21+J24).}
        \label{fig:aggregated_roc_l}
    \end{minipage}
    \label{fig:aggregated_roc_comparison_kl}
\end{figure}

The ROC curves of models in experiment $\mathcal{C}_3$ for test datasets \textit{i}, \textit{j}, \textit{k}, and \textit{l} are shown in Figures \ref{fig:aggregated_roc_i}, \ref{fig:aggregated_roc_j}, \ref{fig:aggregated_roc_k} and \ref{fig:aggregated_roc_l}. Test set \textit{i} is easily classified by most models (near 1.0), while \textit{j} and \textit{k} contain L4 lenses, which are the most challenging samples, as well as the aggregated test set \textit{l}.

\bsp	
\label{lastpage}
\end{document}